\documentclass{article}

    \PassOptionsToPackage{numbers, compress}{natbib}
 \usepackage[preprint]{neurips_2026}

\usepackage[utf8]{inputenc} 
\usepackage[T1]{fontenc}    
\usepackage{hyperref}       
\usepackage{url}            
\usepackage{booktabs}       
\usepackage{amsfonts}       
\usepackage{nicefrac}       
\usepackage{microtype}      
\usepackage{xcolor}         

\usepackage{wrapfig} 
\usepackage{caption}  
\usepackage{amssymb}
\usepackage[most]{tcolorbox}
\newcommand{\appref}[1]{\hyperref[#1]{Appendix~\ref*{#1}}}
\newcommand{\secref}[1]{\hyperref[#1]{Section~\ref*{#1}}}
\tcbuselibrary{listings,breakable}
\usepackage{color}
\usepackage{epsfig}
\usepackage{graphicx}

\usepackage{adjustbox}
\usepackage{array}
\usepackage{booktabs}
\usepackage{colortbl}
\usepackage{wrapfig}
\usepackage{hhline}
\usepackage{multirow}
\usepackage{subcaption}
\usepackage{wrapfig}
\usepackage{floatflt}

\usepackage{bm}
\usepackage{nicefrac}
\usepackage{microtype}

\usepackage{changepage}
\usepackage{extramarks}
\usepackage{fancyhdr}
\usepackage{lastpage}
\usepackage{setspace}
\usepackage{soul}
\usepackage{xspace}

\usepackage{enumerate}
\usepackage{enumitem}  

\usepackage{makecell}

\usepackage{pifont} 

\usepackage{algorithm,algpseudocode}

\usepackage[symbol]{footmisc}
\renewcommand{\thefootnote}{\fnsymbol{footnote}}

\newcolumntype{L}[1]{>{\raggedright\let\newline\\\arraybackslash\hspace{0pt}}m{#1}}
\newcolumntype{C}[1]{>{\centering\let\newline\\\arraybackslash\hspace{0pt}}m{#1}}
\newcolumntype{R}[1]{>{\raggedleft\let\newline\\\arraybackslash\hspace{0pt}}m{#1}}

\newcommand{\ignore}[1]{}

\makeatletter
\DeclareRobustCommand\onedot{\futurelet\@let@token\@onedot}
\def\@onedot{\ifx\@let@token.\else.\null\fi\xspace}

\makeatother

\definecolor{MyDarkBlue}{rgb}{0,0.08,0.8}
\definecolor{MyDarkGreen}{RGB}{45,155,45}
\definecolor{MyDarkRed}{rgb}{0.8,0.02,0.02}
\definecolor{MyOrange}{rgb}{1.0, 0.4, 0.2}
\definecolor{MyPurple}{RGB}{111,0,255}
\definecolor{MyRed}{rgb}{0.8,0.0,0.0}
\definecolor{MyGold}{rgb}{0.75,0.6,0.12}
\definecolor{MyDarkgray}{rgb}{0.66, 0.66, 0.66}

\newcommand{\model}{APIVOT\xspace}

\usepackage{amssymb}
\usepackage{pifont}

\title{APIVOT: Adaptive Planning with Interleaved Vision-Language Thoughts}

\author{%
  Emily Jin
  \quad
  Joy Hsu
  \quad
  Yiqing Xu
  \quad
  Weiyu Liu\textsuperscript{\textdagger}
  \quad
  Nick Haber\textsuperscript{\textdagger}
  \quad
  Jiajun Wu\textsuperscript{\textdagger} \\[4pt]
  Stanford University
}
\begin{document}

\maketitle
\begingroup
\renewcommand{\thefootnote}{\textdagger}
\footnotetext{Equal advising.}
\endgroup

\vspace{-0.2cm}
\begin{abstract}
Long-horizon robot planning requires jointly reasoning over semantic task structure and geometric feasibility. To successfully execute a task, a robot must decompose goals, select task-relevant objects, and sequence actions, while ensuring that plans satisfy spatial constraints such as limited free space and object collisions. In this work, we propose \model, a VLM-based planner that adaptively interleaves language and visual thoughts for long-horizon planning. \model learns to leverage language for semantic reasoning, while using visual thoughts as imagined future states for internal verification of geometric feasibility. On long-horizon kitchen tasks, \model outperforms general-purpose VLMs and prior planning frameworks, achieving the largest gains in spatially constrained settings. We find that \model learns meaningful modality selection behavior, demonstrating that adaptive interleaving of vision-language thoughts improves both planning success and reasoning efficiency. \footnote{Project page: \url{https://emilyzjin.github.io/projects/apivot.html}}
\end{abstract}
\vspace{-0.2cm}

\section{Introduction}
\label{sec:intro}

Long-horizon robot planning requires flexibly interleaving semantic and geometric reasoning. Consider the task: ``store the leftovers in the fridge.'' To successfully achieve this goal, a robot must reason semantically to identify which items need to be stored, select appropriate containers, and determine a sequence of actions that satisfies prerequisites (e.g., the fridge must be open before placing anything inside). At the same time, successful execution depends on geometric constraints, such as whether the leftovers fit inside the selected containers, how the containers should be arranged inside the fridge, and whether any existing objects must be moved aside to create enough free space. These two modes of reasoning are deeply intertwined. A symbolically valid plan may still fail if the containers collide, while geometric decisions made early on can affect which actions remain feasible downstream. 

Existing LLM- and VLM-based planners can reliably produce semantically plausible action sequences, but often struggle when success depends on geometric feasibility~\cite{ahn_as_2022, huang_language_2022, huang_inner_2022, liu_llmp_2023, singh2023progprompt, liang2023code, song2023llm, wu2023embodied}. Prior work addresses this by coupling the model with external motion planners, feasibility checkers, or learned dynamics models~\cite{joublin_copal_2024, yang_guiding_2024, wang_llm3large_2024, skreta_replan_2024, feng_reflective_2025}. However, these systems typically incorporate geometric feedback for replanning, which does not shape the planner's internal reasoning. Instead, we argue that an effective planner should interleave semantic and geometric reasoning itself, using language and vision as complementary modalities. Language is effective for semantic reasoning, such as task decomposition and action selection~\cite{zawalski_robotic_2025}, but it cannot express the geometric structure needed to plan over resulting scene configurations in a compact, precise way. By contrast, visual representations effectively encode spatial layout, object shapes, and remaining free space, making them well-suited for geometric reasoning~\cite{zhao_cot-vla_2025, hu_bagelvla_2026}. Thus, language and vision are complementary modes of reasoning that should be used adaptively based on a task's demands. While recent vision-language-action (VLA) models begin to incorporate visual representations for intermediate reasoning, they apply them uniformly, and a challenge remains in learning when to reason in vision and language ~\cite{zawalski_robotic_2025, zhao_cot-vla_2025, hu_bagelvla_2026, qu_eo1_2025, yuan_fsd_2025, physicalintelligence_pi07_2026}.

\begin{figure}
    \centering
    \includegraphics[width=\linewidth]{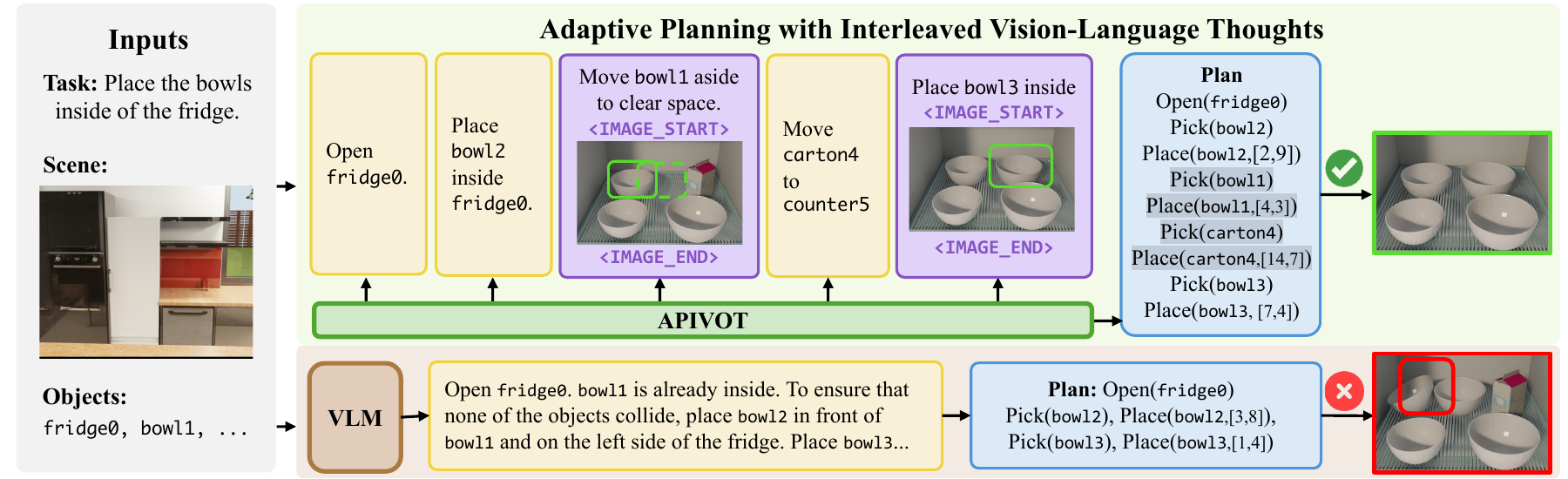}
    \captionsetup{font=small, labelfont=bf}
    \caption{\model (above) plans for long-horizon tasks by interleaving language reasoning with visual thoughts. While a standard VLM planner (below) reasonsing in text may produce a semantically plausible but geometrically infeasible plan, \model{} imagines future states inside its reasoning trace. These visual thoughts reveal potential collisions or constraints before execution, enabling the planner to verify geometric constraints during planning.}
    \label{fig:pull}
    \vspace{-0.9cm}
\end{figure}

To this end, we propose \model, a VLM-based planner that adaptively interleaves language with visual thoughts to perform long-horizon robot planning (\autoref{fig:pull}). Given a task instruction and an image observation of the scene, our model first produces a reasoning trace of interleaved language and visual thoughts, before outputting a sequence of actions for execution. It learns to use language for semantic reasoning about subgoal decomposition and action ordering, and visual thoughts that represent imagined future states for verifying geometric feasibility and guiding downstream planning. Crucially, our model learns \textit{adaptive} modality selection, choosing an appropriate modality at each planning step to jointly perform semantic and geometric reasoning.

\model is trained with supervised fine-tuning (SFT) on reference traces that demonstrate interleaved vision-language thoughts. These traces supervise the model to generate and reason with visual thoughts, while the corresponding latent visual states are aligned with encoded ground-truth images of the scene at different intermediate states. We train \model under a three-stage curriculum that progressively teaches it to plan adaptively. In the first stage, the model learns to reason over provided visual thoughts and extract geometric information from them for planning. In the second stage, it learns to autoregressively generate these visual thoughts itself and optimize them for downstream planning. In the third stage, it learns adaptive modality selection, generating visual thoughts only when they are useful for the current planning step.

We evaluate \model on a suite of long-horizon planning tasks in KitchenWorlds~\cite{yang_guiding_2024}. Across all task families, \model significantly outperforms existing approaches, including general-purpose VLMs, VLM-based planning frameworks, and symbolic planners. Our model achieves an average task success rate of 0.419, demonstrating an 8.1 percentage point improvement over the top-performing baseline. In particular, \model demonstrates the largest advantage in geometrically constrained settings, widening the gap from 7 to 17 points as complexity increases. Beyond improving task success, we find that our method achieves these gains with strong computational efficiency. \model consistently outperforms language-only VLMs across various reasoning budgets. This suggests that visual thoughts provide a compact, expressive representation for planning, capturing spatial information that would otherwise be verbose or difficult to specify in language alone. We further analyze the modality selection behavior learned by \model. While using visual thoughts at every step provides an upper bound on performance, \model retains $91\%$ of this performance while substantially reducing token usage by $39\%$. This efficiency gain is consistent with our finding that \model selectively generates visual thoughts on steps requiring geometric precision. Together, these results suggest that \model learns meaningful modality selection for downstream planning.

In summary, our contributions are the following:
\begin{itemize}[left=0.5em, labelsep=0.5em, itemsep=0.1em]
\vspace{-0.2cm}
    \item We introduce \model, a VLM-based planner that interleaves language for semantic reasoning and visual thoughts as imagined future states for internal geometric reasoning. 
    \item We propose a three-stage training curriculum that progressively teaches the model to reason over, generate, and adaptively use visual thoughts during long-horizon planning. 
    \item We demonstrate that our method improves both task success and efficiency across a diverse set of ecological, long-horizon planning tasks, outperforming general-purpose VLMs and planning frameworks while reducing reasoning cost through adaptive modality selection.
\end{itemize} 
\section{Related Work}
\label{sec:related_work}

\vspace{-0.2cm}
\paragraph{Reasoning for long-horizon robotic manipulation.}

Long-horizon robotic manipulation requires reasoning over both symbolic task structure and geometric feasibility. Existing approaches can be broadly grouped into three families. Classical task and motion planning methods combine symbolic search with geometric feasibility checking or sampling~\cite{garrett_integrated_2021, garrett_pddlstream_2020, pddl1998, helmert_fast_2006}. While principled, they require hand-specified symbolic abstractions and domain models. By contrast, LLM- and VLM-based planners leverage pre-trained models to decompose natural-language instructions into action sequences, programs, or high-level plans~\cite{ahn_as_2022,huang_language_2022, huang_inner_2022, liu_llmp_2023, singh2023progprompt, liang2023code, song2023llm, wu2023embodied}. These approaches offer greater flexibility, but often reason primarily at the semantic level, producing plans that are plausible in language but infeasible geometrically. A third line of work improves the geometric grounding of LLM- and VLM-based planners by integrating them with external planners, verifiers, or learned dynamics models~\cite{joublin_copal_2024, yang_guiding_2024, wang_llm3large_2024,  skreta_replan_2024, feng_reflective_2025, xu2024set, yan_using_2026}. The external modules provide geometric feedback through feasibility checks and replanning, improving execution but leaving geometry outside of the model's own reasoning process. Our work belongs to the broader class of VLM-based planners, but integrates geometric reasoning directly into planning. \model generates visual thoughts within its own reasoning trace, producing imagined future states to condition subsequent decisions before the plan is produced. In doing so, geometric reasoning is an internal part of planning rather than a post-hoc verification signal. 

\vspace{-0.2cm}
\paragraph{Multimodal reasoning and vision-language-action models.}

While prior planning methods often introduce geometry through external verifiers or simulators, an alternative is for the planner to perform multimodal reasoning itself. Recent unified multimodal models learn shared frameworks for visual understanding and generation~\cite{bai_qwen3-vl_2025, noauthor_gemini_nodate, deng_bagel_2025, xie_showo2_2025, liu_tuna_2025, liu_tuna2_2026, tong_beyond_2026}. These models support general-purpose visual generation, but do not address how generation should guide long-horizon decisions. A related line of work studies multimodal chain-of-thought or latent visual reasoning, where models reason with intermediate visual representations instead of purely in text~\cite{su_thinking_2025, li_zebra-cot_2025, li_imagine_2025, qin_covt_2025, yang_machine_2025, gu_thinkmorph_2026, ma_latte_2025, wang_pixel_2025, ray_mulltokens_2025, lu_thinking_2026}. These approaches improve visual and spatial reasoning, but largely operate on the current observation through attention, visual marking, perception refinement, or static problem solving~\cite{li_imagine_2025, qin_covt_2025, yang_machine_2025, gu_thinkmorph_2026}. However, robot planning requires reasoning beyond the current image to planning how the scene will evolve over multiple actions. A third line of work brings multimodal reasoning into robotics through vision-language-action models and embodied foundation models~\cite{zawalski_robotic_2025, zhao_cot-vla_2025, hu_bagelvla_2026, qu_eo1_2025, yuan_fsd_2025, physicalintelligence_pi07_2026}. These methods use visual intermediates, spatial representations, or multimodal context to improve action generation, but primarily target policy execution rather than high-level semantic planning. In contrast, \model{} uses visual thoughts within a high-level planning trace, learning to use vision and language adaptively. 

\vspace{-0.2cm}
\paragraph{Adaptive and efficient multimodal reasoning.}

A separate line of work aims to improve models' reasoning efficiency by asking \emph{how much} computation a model should allocate. Some methods allocate different thinking budgets, compress reasoning into latent tokens, or decide when explicit chain-of-thought is needed~\cite{ray_mulltokens_2025, lou_adacot_2025}. In multimodal settings, related methods reduce the cost of visual reasoning by making the visual representation more compact, using latent visual tokens, specialist visual modules, or explicit primitives such as points and bounding boxes~\cite{qin_covt_2025, yang_machine_2025, ma_latte_2025, wang_pixel_2025, lu_thinking_2026}. In embodied settings, prior work improves reasoning efficiency through lightweight training strategies or by interleaving intermediate reasoning with action prediction~\cite{zawalski_robotic_2025, chen_ecotlite_2025}. In general, these approaches improve efficiency by making reasoning cheaper or more selectively allocated.

In robot planning, however, efficiency can also depend on \emph{which representation} is appropriate for a given decision. Language is efficient and expressive for semantic decisions such as goal decomposition, object selection, and preconditions. Visual thoughts are more appropriate for geometric reasoning about fitting objects into constrained spaces or evaluating remaining free space, but are also more expensive~\cite{shi_large_2023, liu_lost_2023, wu_how_2024}. \model{} therefore learns adaptive modality selection. It chooses between language and visual thoughts according to the structure of the planning decision. This makes adaptivity task-grounded, improving performance while reducing unnecessary visual reasoning.\section{Vision-Language Adaptive Planning}
\label{sec:method}

\begin{figure}
    \centering
    \includegraphics[width=\linewidth]{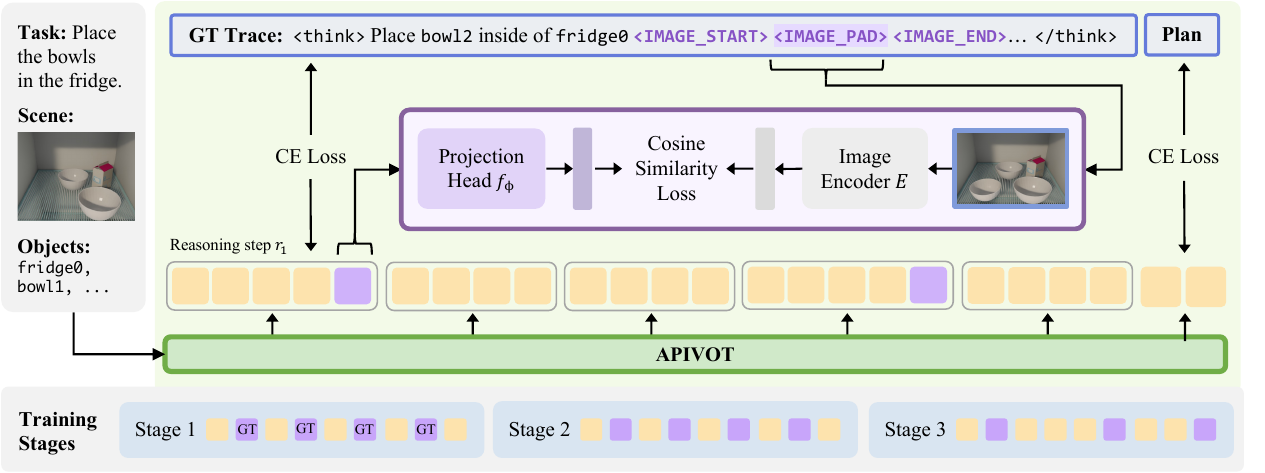}
    \captionsetup{labelfont=bf}
    \caption{\model produces a reasoning trace of interleaved language tokens (yellow) and visual-thoughts (purple) followed by a plan, and is trained with a three-stage SFT curriculum. In all stages, we apply teacher-forced SFT to a reference reasoning trace and plan (blue border). Stage 1 teaches the model to plan with ground-truth visual thoughts, provided at every reasoning step; Stage 2 trains it to generate visual thoughts itself by aligning them to encoded ground-truth future observations via a cosine similarity loss; and Stage 3 omits visual thoughts for some reference steps, teaching the model to plan adaptively with visual thoughts.}
    \label{fig:system}
    \vspace{-0.4cm}
\end{figure}

In this section, we present \model, a VLM-based planner that learns to plan by interleaving language with visual thoughts within its reasoning trace, as illustrated in \autoref{fig:system}. We begin by giving an overview of the problem formulation and describing the overall planning pipeline (\secref{sec:method:overview}). After, we introduce our supervised fine-tuning setup (\secref{sec:method:supervision}). Finally, we present the training curriculum that teaches the model to adaptively generate visual thoughts alongside language and leverage them for planning (\secref{sec:method:training}).

\vspace{-0.2cm}
\paragraph{Problem formulation.}
\label{sec:method:problem_formulation}
We consider goal-conditioned planning in a Semi-Markov Decision Process (semi-MDP) with parameterized
skills~\citep{sutton1999between,garrett_integrated_2021}. The environment is a
tuple $(\mathcal{S}, \mathcal{A}, \mathcal{T})$, where $\mathcal{S}$ is
the state space, $\mathcal{A}$ is a finite library of parameterized
skills, and $\mathcal{T}: \mathcal{S} \times \mathcal{A} \rightarrow
\mathcal{S}$ is the transition function induced by skill execution. Each
skill $a \in \mathcal{A}$ takes discrete object arguments together with
optional continuous parameters. In this work, we instantiate
$\mathcal{A}$ with three skills: \texttt{open(obj)}, \texttt{pick(obj)},
and \texttt{place(obj,\,target,\,$u$,\,$v$)}, where $(u,v)$ is an
image-space placement coordinate that is lifted to a world-frame pose
using the observed depth image and the known camera extrinsics. An
episode is specified by an initial state $s_0$ and a natural-language
goal $g$, and terminates when $g$ is achieved or a timeout is reached.
We assume access to a dataset $\mathcal{D} = \{\tau_i\}_{i=1}^{N}$ of
successful demonstrations, where each trajectory $\tau_i = ((s^t_i,
a^t_i))_t$ is annotated with the skill $a^t_i$ applied at each step.

At each timestep $t$, the planner observes an image observation $I_t$
of the current state $s_t$ together with a list of detected objects
$\mathcal{O}_t = \{(o_i, [x^t_i, y^t_i])\}$, where $o_i$ is an object
instance and $[x^t_i, y^t_i]$ is its image-space location. Detected
objects are provided as input to isolate planning from perception,
allowing the model to focus on reasoning and decision-making. We
execute plans in a closed-loop, receding-horizon setup~\citep{mayne2000constrained}: given
$(I_t, \mathcal{O}_t)$ and the goal $g$, the planner outputs a remaining
plan $(a_t^1, \ldots, a_t^N)$ of skill calls; the first skill $a_t^1$ is
applied to the environment, the state transitions to $s_{t+1}$, and the
planner is re-invoked with the updated observation. This setup
encourages the model to reason over the full task while grounding each decision in the observed state after prior
actions.

\subsection{Planner Overview}
\label{sec:method:overview}

At inference, \model generates a \emph{multimodal reasoning trace} and a \emph{plan}. The trace interleaves the model's reasoning about task decomposition, action dependencies, and constraints, with \emph{visual thoughts}. The plan specifies a sequence of actions in the planner action space $\mathcal{A}_{\mathrm{plan}}$, including both discrete primitives and placement parameters.

\vspace{-0.2cm}
\paragraph{Multimodal reasoning trace.} \model generates a reasoning trace $R := r_1\oplus \ldots \oplus r_m$, where $\oplus$ denotes concatenation and each $r_j$ is a reasoning step associated with an intermediate subgoal $\rho_j$. As shown in \autoref{fig:system}, each step consists of text tokens $t_j$ and may optionally include visual tokens $v_j$. The text tokens describe the subgoal $\rho_j$, including the object to interact with, its role in advancing the task, and the relevant semantic or geometric constraints. When visualizing the future scene is useful for verifying subgoal feasibility, the model generates a span of visual tokens $v_j$, which we call a \textit{visual thought}. Otherwise, $v_j=\varnothing$. This formulation supports adaptive multimodal reasoning. The model can rely on language for semantic reasoning and introduce visual thoughts only when needed.

\vspace{-0.2cm}
\paragraph{Visual thought.} A visual thought is instantiated in the trace as a fixed-length span of $K$ \emph{visual tokens}, delimited by special start and end tokens: \texttt{<|image\_start|><|image\_pad|>$^K$<|image\_end|>}. These discrete tokens indicate where the visual thought occurs in the autoregressive sequence, while the decoder hidden states at the $K$ visual tokens encode its visual content. We denote these hidden states as the \emph{latent visual state} $H_j \in \mathbb{R}^{K \times d}$, where $d$ is the VLM's hidden dimension. Since $H_j$ remains in the autoregressive context, subsequent tokens can attend to this predicted future-state representation during downstream reasoning and plan generation. 

\vspace{-0.2cm}
\paragraph{Action plan.} Conditioned on the reasoning trace, the model outputs a plan $(a_1,\ldots,a_n)$, where each $a_i\in\mathcal{A}_{\mathrm{plan}}$ specifies a discrete primitive such as \texttt{open}, \texttt{pick}, or \texttt{place}, along with any required placement parameters. Conceptually, the language components of the trace support discrete action selection, while the visual thoughts encode spatial context for grounding placement parameters. 

\subsection{Supervision from Reference Outputs}
\label{sec:method:supervision}

We initialize \model from a pretrained VLM that can process visual inputs and conduct text-based reasoning, but cannot generate or use visual thoughts for planning. To plan effectively, we perform supervised fine-tuning (SFT) on reference outputs derived from successful demonstrations. Each reference output consists of a multimodal reasoning trace and plan, paired with ground-truth images. 

\vspace{-0.2cm}
\paragraph{Reference output construction.} 
We construct reference outputs by decomposing each demonstration $\tau \in \mathcal{D}$ into a sequence of subgoals. Consecutive \texttt{pick}/\texttt{place} skill calls are paired, and each subgoal is annotated with three labels: a constraint type, a modality label, and a purpose. The constraint type captures what kind of reasoning is required, while the modality label specifies whether the reference trace should solve it with text alone or include a visual thought. Concretely, we define three constraint types: a \emph{symbolic precondition}, where the decision is independent of continuous parameters (e.g., opening the fridge); a \emph{current geometric constraint}, where the parameter is restricted by the present scene (e.g., a large bowl in a near-full fridge); and \emph{future geometric feasibility}, where the parameter is locally unconstrained but determines whether later subgoals remain feasible (e.g., placing a small bowl so a larger one can fit later). The latter two involve continuous parameters and include cases where greedy refinement of a high-level plan can fail and downward refinability can break down~\cite{garrett_integrated_2021}. We compute the constraint type from geometric metrics over depth and camera extrinsics, including surface areas, object footprints, and free-space margins, and map it to a modality label by thresholding the resulting tightness: symbolic-precondition steps are text-only, and geometric steps receive a visual thought once tightness exceeds the threshold. We additionally tag each subgoal with a purpose, such as advancing the task, satisfying a precondition, or clearing an
obstacle. 

We then assemble the annotated subgoals into a structured skeleton and prompt a high-capacity language model to expand it into a natural-language reasoning trace. We instruct it to preserve the skeleton's structure while adding connective reasoning about task decomposition, subgoal selection, and progress toward the goal. Conditioning expansion on the subgoals and annotations grounds the trace in the physical constraints relevant to planning rather than generic rationalization, while still matching the VLM's natural reasoning style. For every reasoning step labeled with a visual thought, we insert the discrete visual token sequence after the text component of the step. We also attach a target future RGB observation of the environment after the corresponding subgoal is achieved. This image provides supervision for aligning the visual thought to the ground-truth future state at that point in the trace. Using this procedure, we construct SFT datasets with different modality-labeling strategies to induce different behaviors, such as using visual thoughts at every step or adaptively. See \appref{app:data:pipeline} for more details and examples of reasoning traces.

\vspace{-0.2cm}
\paragraph{Visual alignment.} 
To supervise the VLM's latent visual states, we align them to target features obtained by encoding the corresponding ground-truth images with a frozen vision encoder $E$. We use the pretrained VLM's own vision encoder to encourage the visual thoughts to lie within the model's visual feature space. Let $I^{\mathrm{gt}}_j$ be the reference RGB image for the $j$-th visual thought. We encode $I^{\mathrm{gt}}_j$ with $E$ and average-pool the resulting patch features along the patch dimension into $K$ target features: $Z^{\mathrm{gt}}_j = \mathrm{AvgPool}\bigl(E(I^{\mathrm{gt}}_j)\bigr) \in \mathbb{R}^{K \times d_{E}},$ where $d_{E}$ is the feature dimension of $E$. We then project each latent visual state $H_j$ into the encoder feature space using a learned projection head $f_\phi: \mathbb{R}^{d} \rightarrow \mathbb{R}^{d_{E}}$ to obtain $Z_j = f_\phi(H_j)$. Finally, we apply a cosine similarity loss over the projected latent states: $\mathcal{L}_{\mathrm{vis}}
    =
    \frac{1}{|\mathcal{B}_{\mathrm{vis}}|}
    \sum_{j\in \mathcal{B}_{\mathrm{vis}}} 
        1 -\cos(Z_{j}, Z^{\mathrm{gt}}_{j})
    ,$
where $\mathcal{B}_{\mathrm{vis}}$ is the set of visual thoughts in the batch.  

\vspace{-0.2cm}
\paragraph{Training objective.} Across stages, we optimize a training objective with three components: $\mathcal{L}_{\mathrm{total}} = \mathcal{L}_{\mathrm{CE, plan}} + \lambda_{\mathrm{CE,trace}}\mathcal{L}_{\mathrm{CE, trace}} + \lambda_{\mathrm{vis}}\mathcal{L}_{\mathrm{vis}},$
where $\lambda_{\mathrm{CE,trace}}$ and $\lambda_{\mathrm{vis}}$ are hyperparameters that control the weights of the loss components.
Here, $\mathcal{L}_{\mathrm{CE,trace}}$ and $\mathcal{L}_{\mathrm{CE,plan}}$ are cross-entropy losses computed over tokens in the multimodal reasoning trace and plan, respectively. The $\mathcal{L}_{\mathrm{CE,trace}}$ loss supervises the structure and content of the reasoning trace, including the syntax of visual thoughts, while the $\mathcal{L}_{\mathrm{CE,plan}}$ trains the model to generate the successful plan conditioned on the reasoning trace. Meanwhile, the visual loss $\mathcal{L}_{\mathrm{vis}}$ aligns the latent visual states with the target scene content. 

\subsection{Training Curriculum}
\label{sec:method:training}
Our three-stage training curriculum teaches the model how to: 1) \emph{use} visual thoughts as intermediate context for planning, 2) \emph{generate} them internally, and 3) decide \emph{when} they are useful. 

 \vspace{-0.2cm}
\paragraph{Stage 1: Visual thought comprehension.} The first stage teaches the model to use visual thoughts to guide planning. To isolate this capability from visual thought generation and modality selection, we train on traces where every reasoning step includes a visual thought, and provide the target latent visual states directly to the model. In particular, we inject the corresponding ground-truth visual features into the decoder input at all visual token positions, providing context for generating subsequent reasoning and actions. As the latent visual states are provided, we disable the visual alignment loss and supervise only the discrete output tokens. In doing so, the model learns to generate the reference reasoning text and effectively ground planning in the provided visual states. 

\vspace{-0.2cm}
\paragraph{Stage 2: Visual thought generation.} The second stage teaches the model to generate the visual thoughts itself. As in Stage 1, every reasoning step includes a visual thought, so that the model learns the mechanics of visual thought generation without having to decide when they are necessary. Instead of injecting the ground truth embeddings, the model must now generate its own latent visual states at the visual placeholder positions. We optimize all loss components by setting $\lambda_{\mathrm{vis}}>0$, encouraging the model to generate the visual tokens and the latent visual state to encode the target scene context.  

 \vspace{-0.2cm}
\paragraph{Stage 3: Adaptive modality selection.} The final stage teaches the model to decide the appropriate reasoning modality for each subgoal and generate visual thoughts only when useful. We train on reasoning traces with adaptive modality selection, where visual thoughts are generated only for subgoals that benefit from geometric grounding. We assign modality labels based on simulator-derived geometric heuristics that identify subgoals constrained by limited free space, collision-sensitive placements, or downstream feasibility dependencies. In this stage, we include all training loss components. The trace CE loss supervises when to use visual thoughts, while the visual alignment and plan CE losses preserve latent visual state generation and downstream planning behaviors, respectively. Together, these losses encourage the model to retain the benefits of planning with visual thoughts, while learning to emit them selectively to reduce unnecessary reasoning cost.

\section{Experiments}
\label{sec:experiments}
Our experiments address two primary questions. First, we evaluate whether \model improves long-horizon planning performance, and further analyze when these gains arise. Second, we understand the extent to which \model demonstrates meaningful adaptive modality selection behavior.

\subsection{Experimental Setup}

\vspace{-0.2cm}
\paragraph{Tasks.} We design a suite of long-horizon kitchen tasks that require reasoning jointly about semantic and geometric constraints. The tasks systematically vary semantic goals, geometric constraints, and their compositions, enabling controlled evaluation in settings that demand precise geometric reasoning. The suite consists of three task families: \textsc{containment}, \textsc{sorting}, and \textsc{storing leftovers}. \textsc{Containment} involves placing objects into a constrained storage space, requiring reasoning about semantic preconditions and limited free space. \textsc{Sorting} involves assigning objects to containers by type, often under capacity constraints that restrict feasible assignments. \textsc{Storing Leftovers} composes these into a longer-horizon task, with sorting followed by placement into a shared constrained space. We train on \textsc{Containment} and \textsc{Sorting}, and hold-out \textsc{storing leftovers} to evaluate transfer and compositional generalization. See \appref{app:data:task_families} for examples. 

\begin{figure}
    \centering
    \includegraphics[width=\linewidth]{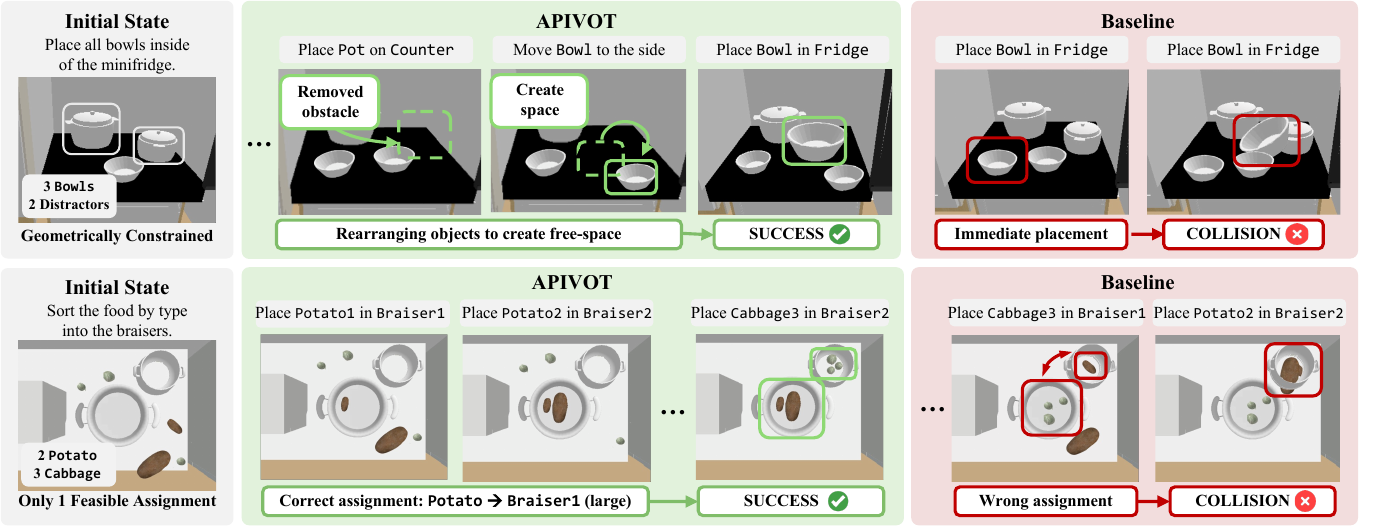}
    \captionsetup{labelfont=bf}
    \caption{We show execution traces of \model and the top-performing baseline on each task: \textsc{Containment} (top) and \textsc{Sorting} (bottom).}
    \label{fig:qualitative_examples}
    \vspace{-0.7cm}
\end{figure}

We generate all data in the KitchenWorlds~\cite{yang_guiding_2024} simulator. Our pipeline samples diverse task instances with varying scene layouts and object configurations, producing settings that range from simple cases with ample free-space to harder ones requiring opening doors, removing obstacles, or carefully arranging objects under tight space. For each task instance, we use PDDLStream~\cite{garrett_pddlstream_2020} to generate a successful trajectory and synthesize the corresponding reference output. We train on 2{,}000 examples per training task, and evaluate on held-out datasets for each training task as well as the held-out task, each with 100 examples. Additional dataset details are provided in \appref{app:data:pipeline}.

We evaluate methods in a receding horizon setup with access to a shared library of primitive actions. Our primary metric is \emph{task success}, defined as the fraction of episodes that succeed. To measure reasoning cost for VLM-based methods, we report \emph{token usage}, computed as the number of tokens in the reasoning trace. For extended thinking VLMs, we include the internal reasoning tokens.

\vspace{-0.2cm}
\paragraph{Model implementation.} \model is initialized from Qwen3-VL-8B-Instruct~\citep{bai_qwen3-vl_2025} and finetuned with LoRA applied to the attention and MLP projections. Full details are provided in \autoref{app:model}. 

\vspace{-0.2cm}
\paragraph{Baselines.} We compare \model with three categories of planners: general-purpose VLMs, VLM-based planners, and symbolic planners. For general-purpose VLMs, we use Gemini-3.1-Pro, Gemini-ER-1.5~\cite{noauthor_gemini_nodate}, Qwen3-VL-8B-Instruct (\model's base model), and Qwen3-VL-8B-Thinking. Extended thinking is enabled for Gemini-3.1-Pro, Gemini-ER-1.5, and Qwen3-VL-8B-Thinking. In addition, we choose VLM-TAMP~\cite{yang_guiding_2024} and Reflect-VLM~\cite{feng_reflective_2025} as representative VLM-based planners and FastDownward~\cite{helmert_fast_2006} as the symbolic planner. Since FastDownward assumes access to symbolic states, we provide it with symbolic states inferred from visual input and language following BLADE~\cite{liu2024learning}, to ensure a fair comparison. More implementation details are in \autoref{app:baselines}.

\subsection{Task Planning Performance}
\label{sec:results:performance}

\begin{wraptable}{r}{0.55\linewidth}
\vspace{-1cm}
\captionsetup{labelfont=bf}
  \caption{Planning performance across task families.}
  \label{tab:main_perf}
  \centering
  \resizebox{\linewidth}{!}{
  \begin{tabular}{l c c c c}
    \toprule
    Model
    & Avg 
    & \textsc{Contain} 
    & \textsc{Sort} 
    & \textsc{Store} \\
    \midrule
    \multicolumn{5}{l}{\textit{VLM Baselines}} \\
    Gemini-3.1-Pro        & 0.245 & 0.281 & 0.241 & 0.213 \\
    Gemini-ER-1.5         & 0.338 & 0.364 & 0.339 & 0.311 \\
    Qwen3-VL-8B-Instruct  & 0.188 & 0.218 & 0.183 & 0.162 \\
    Qwen3-VL-8B-Thinking  & 0.232 & 0.259 & 0.234 & 0.204 \\
    \midrule
    \multicolumn{5}{l}{\textit{Planning Baselines}} \\
    FastDownward  & 0.272 & 0.331 & 0.247 & 0.238 \\
    Reflect-VLM    & 0.292 & 0.348 & 0.272 & 0.257 \\
    VLM-TAMP      & 0.329 & 0.370 & 0.311 & 0.307 \\
    \midrule
    \textbf{\model (Ours)} 
                 & \textbf{0.419}
                 & \textbf{0.472}
                 & \textbf{0.421}
                 & \textbf{0.365} \\
    \bottomrule
  \end{tabular}
  }
  \vspace{-0.3cm}
\end{wraptable}

\vspace{-0.2cm}
\paragraph{Overall performance.} In \autoref{tab:main_perf} we compare \model against prior work across both training task families and a held-out task family. We split prior methods into general-purpose VLM baselines and planning baselines. \model achieves the highest average task success rate of 0.419. Compared to the top-performing VLM baseline, Gemini-ER-1.5, \model demonstrates an 8.1 percentage point improvement, suggesting that stronger pretrained language-based reasoning is insufficient for these tasks. Compared to the strongest planning baseline, VLM-TAMP, \model improves success by 9.0 points. We hypothesize that these gains come from integrating semantic and geometric reasoning while planning, allowing spatial constraints to shape downstream decisions. These gains are consistent across both training tasks, with \model outperforming prior work by 10.2 points on \textsc{Containment} and 8.2 points on \textsc{Sorting}. Notably, \model also achieves a 5.4 point improvement on \textsc{Storing Leftovers}, a held-out compositional task that combines the two training tasks. This result suggests that \model learns a planning strategy that transfers to novel compositions of semantic and geometric constraints. In \appref{app:results_ood_gen}, we evaluate on out-of-distribution settings with longer task horizons and increased complexity, finding that our model remains competitive. 

\begin{wrapfigure}{r}{0.45\linewidth}
\vspace{-0.4cm}
  \centering
  \includegraphics[width=1.0\linewidth]{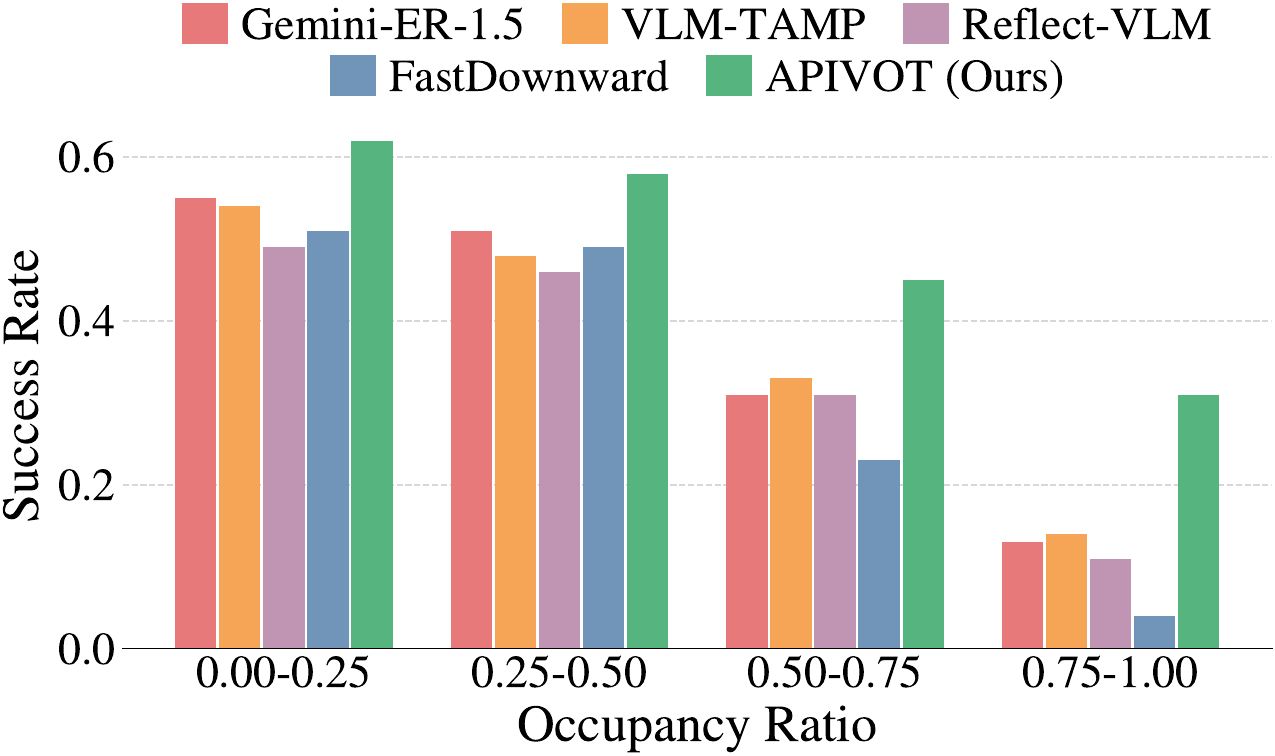} 
  \captionsetup{labelfont=bf}
  \caption{Success rate by \emph{occupancy ratio}.}
  \label{fig:perf_breakdown}
  \vspace{-0.45cm}
\end{wrapfigure}

\vspace{-0.2cm}
\paragraph{Performance under increasing geometric complexity.} To understand where these gains arise, \autoref{fig:perf_breakdown} breaks down task success as a function of geometric complexity. Since all task families require placing objects into target regions, we measure this using an \emph{occupancy ratio}, defined as the total area of task-relevant objects divided by the available target area. 
\begin{wrapfigure}{r}{0.86\linewidth}
\vspace{-0.4cm}
     \centering
    \includegraphics[width=\linewidth]{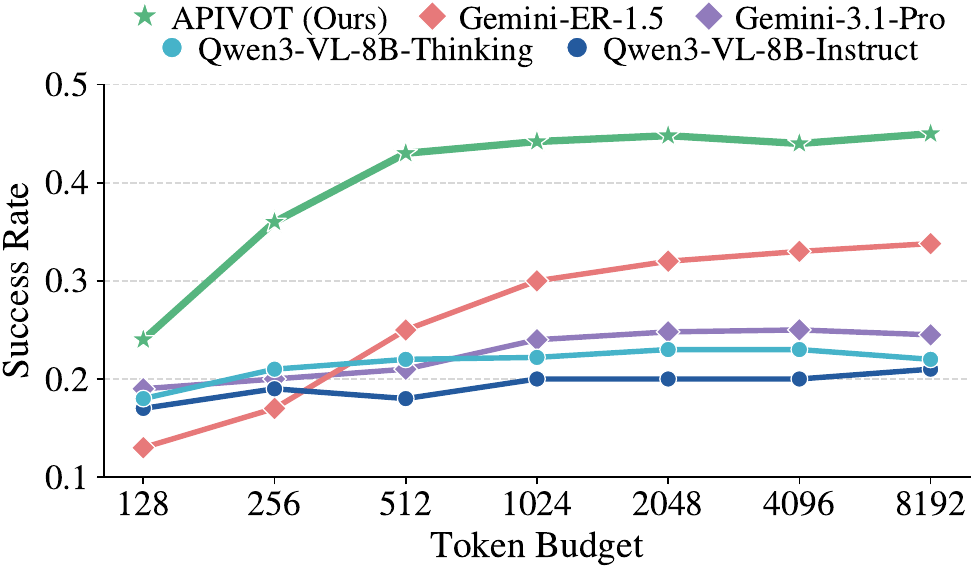}
    \captionsetup{labelfont=bf}
    \vspace{-0.4cm}
    \caption{Success rates of VLMs at different token budgets.}
    \label{fig:pareto_thinking_budget}
    \vspace{-0.2cm}
\end{wrapfigure}

We compare \model against Gemini-ER-1.5 as the strongest representative VLM baseline, as well as all planning baselines. In low-occupancy settings, tasks are under-constrained and all methods achieve high success. As the occupancy ratio increases, however, tasks require more precise placements, and performance decreases across all approaches. Notably, while baseline performance drops significantly under these conditions, \model remains relatively robust, widening the gap from 0.07 to 0.17 points as occupancy increases. \autoref{fig:qualitative_examples} illustrates this advantage: in the \textsc{Containment} task, \model anticipates that the final bowl cannot be placed without more free space. It therefore rearranges existing objects first, enabling success. In comparison, Gemini-ER-1.5 outputs a symbolically valid plan 
but places objects directly, causing a collision. See \appref{app:qualitative_examples} for more examples. 

\vspace{-0.2cm}
\paragraph{Performance under reasoning budgets.} 
To assess whether visual thoughts encode spatial information more efficiently than language, we compare \model against VLM baselines under matched maximum reasoning-token budgets (\autoref{fig:pareto_thinking_budget}). \model consistently outperforms all baselines across these settings, with the largest gains at low- and medium- token budgets. While increasing the token allocation improves some reasoning-heavy baselines, their performance remains below that of \model even at much larger scales. This suggests that visual thoughts provide a more token-efficient representation for planning. Instead of relying on long language traces to describe spatial configurations, \model encodes this information directly and compactly with visual thoughts.

\vspace{-0.2cm}
\paragraph{Role of visual thoughts.}  
To isolate the contribution of visual thoughts, we train a text-only variant of our method using standard SFT, where the reasoning traces only use language. This ablation controls for whether improvements come from domain-specific SFT alone or from the ability to represent intermediate future states visually. Across all tasks, the base model achieves an average success rate of $0.18$, while the text-only finetuned variant shows only limited improvement, reaching $0.24$ (\appref{app:ablations:text_finetune}). In contrast, \model achieves $0.42$, suggesting that while domain-specific SFT is beneficial, the larger gains of \model come from its use of visual thoughts. Additional reasoning trace modality ablations in \appref{app:ablations/reasoning_modality} further disentangle the effects of modality choice. 

\subsection{Adaptive Modality Selection Behavior}

\begin{wrapfigure}{r}{0.45\linewidth}
\vspace{-1cm}
  \centering
  \includegraphics[width=\linewidth]{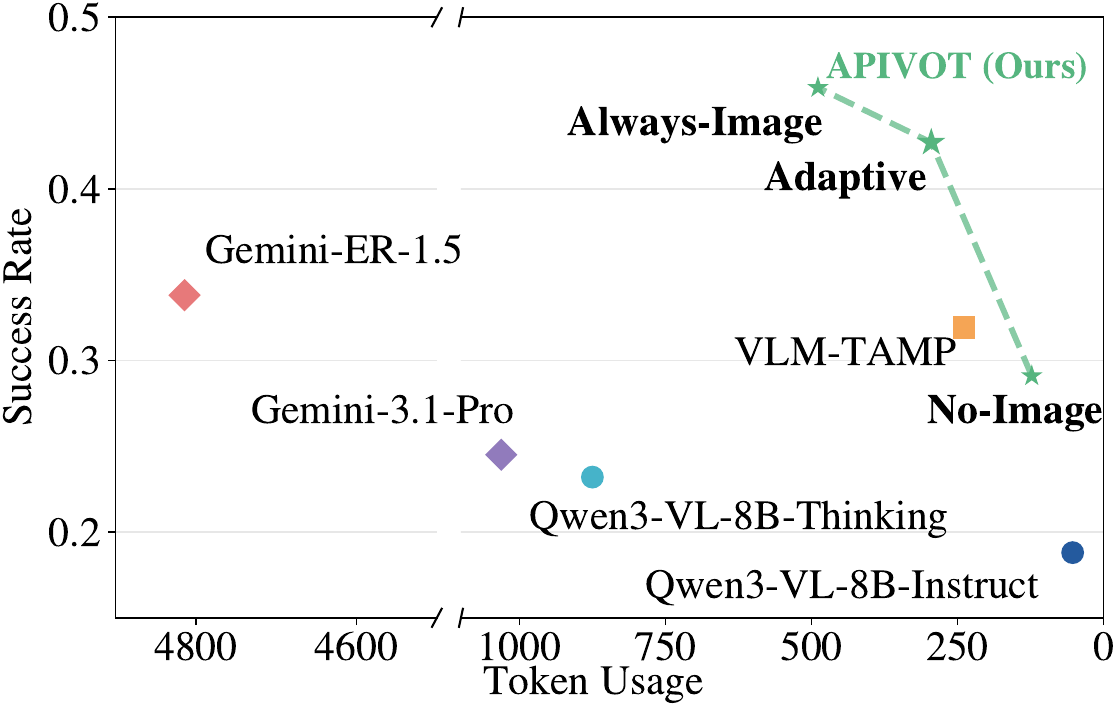}
    \vspace{-0.45cm}
    \captionsetup{labelfont=bf}
  \caption{Perf.-efficiency trade-off.}
  \label{fig:pareto_tokens}
  \vspace{-0.2cm}
\end{wrapfigure}

\vspace{-0.2cm}
\paragraph{Effective modality selection.} 
To evaluate the benefit of adaptive modality selection, we compare our method (\textit{Adaptive}) against two inference-time ablations: text-only reasoning (\textit{No-Image}) and visual thoughts at every step (\textit{Always-Image}). 
As shown in \autoref{fig:pareto_tokens}, \textit{Always-Image} achieves the highest task success but incurs higher token usage from using visual thoughts at every step. In contrast, \model retains $91\%$ of \textit{Always-Image}'s performance while substantially reducing token usage by $39\%$, placing it at a favorable point on the performance-efficiency frontier. Compared to baselines, \model also lies in the high-success, low-cost region of the tradeoff curve. By using visual thoughts selectively, \model avoids unnecessary reasoning while maintaining their benefits for planning. 

\vspace{-0.2cm}
\paragraph{Constraint-aware modality selection.} 
\begin{wrapfigure}{r}{0.45\linewidth}
  \vspace{-1em}
  \centering
  \includegraphics[width=\linewidth]{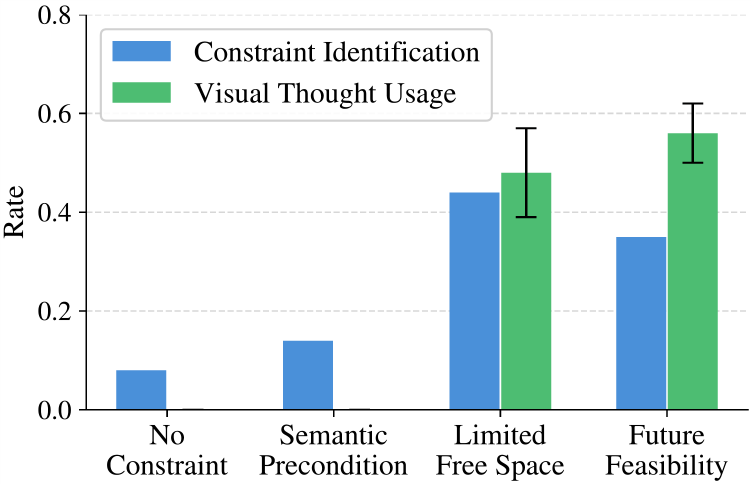}
  \captionsetup{labelfont=bf}
  \caption{Visual thought usage by constraint.}
  \label{fig:image_usage_by_constraint}
  \vspace{-1em}
\end{wrapfigure}

\begin{figure}
  \centering
  \includegraphics[width=\linewidth]{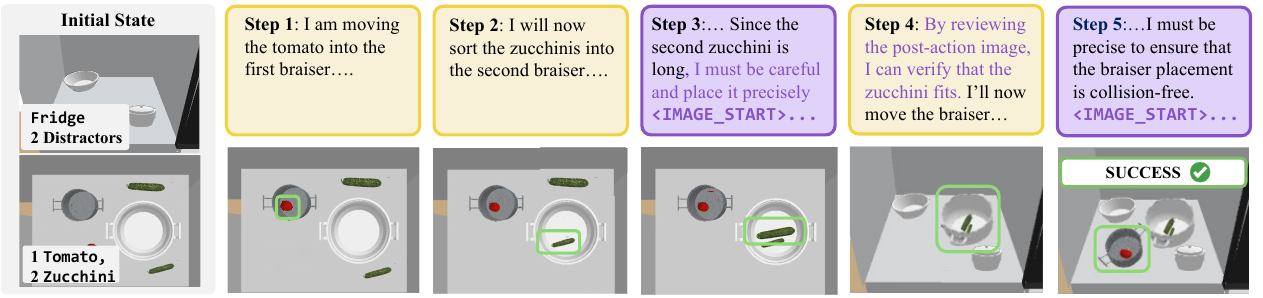}
  \captionsetup{labelfont=bf}
  \caption{Adaptive reasoning behavior on a held-out \textsc{Storing Leftovers} task. APIVOT reasons in language for semantic decisions such as food-to-container assignment, while selectively generating visual thoughts for geometry-sensitive steps, such as collision-free zucchini placement and final fridge storage. This shows that APIVOT invokes visual reasoning when spatial precision is needed, rather than using visual thoughts uniformly.}
  \label{fig:adpative_example}
      \vspace{-0.6cm}
\end{figure}

Beyond its effect on efficiency, we examine whether \model's adaptive use of visual thoughts reflects meaningful modality-selection behavior. We analyze our model's reasoning traces to see whether it uses visual thoughts systematically based on the constraint it identifies at each reasoning step. For different constraint types, \autoref{fig:image_usage_by_constraint} reports how often \model identifies a step as involving that constraint (blue), and among those steps, how often it generates a visual thought (green). When our model determines that a step does not require geometric reasoning (e.g., unconstrained, semantic precondition), it does not produce visual thoughts. On the other hand, it uses visual thoughts for $48.4\%$ of steps involving limited free space and $56.2\%$ of steps involving downstream feasibility constraints. We see an example of such adaptive reasoning in \autoref{fig:adpative_example}. This suggests that \model has learned to use visual thoughts selectively based on when precision matters, rather than applying them uniformly. 
\section{Conclusion}
\vspace{-0.1cm}
\label{sec:conclusion}

We propose \model, a VLM-based planner that learns to leverage interleaved language and visual thoughts for long-horizon task planning. Our results show that incorporating visual thoughts improves planning performance, with the largest gains in spatially constrained settings. Beyond improving success, our model learns meaningful modality selection behavior, choosing to reason in language for semantic task structure and in vision when geometric precision is useful. With this adaptive strategy, \model achieves strong performance gains while maintaining efficiency, demonstrating that learning to reason in the right modality is critical to effective multimodal planning.

\vspace{-0.2cm}
\paragraph{Limitations.} First, our experiments are conducted in the KitchenWorlds simulator. While the task suite captures an ecological range of symbolic and geometric planning challenges, evaluating it in more visually realistic simulators or real-world settings remains an important next step. Second, our learned latent visual representations are trained within our task distribution and may not capture the full variability of real-world geometry and appearance. Scaling visual pretraining to larger, real-world datasets or leveraging advances in unified text-and-image generation models could further improve performance and generalization. Third, adaptive modality selection is learned via supervised finetuning rather than being explicitly optimized for task success. While our model already exhibits semantically meaningful image use, reinforcement learning or test-time adaptation to optimize abstraction selection for task success could yield a more robust policy. Finally, our framework reasons over text and latent images. Extending it to include additional representations (e.g. points or bounding boxes) is a natural direction for enabling it to reason at more levels of abstraction, improving planning performance and efficiency.

\subsubsection*{Acknowledgments}
This paper is based upon work supported by the Air Force Office of Scientific Research under award number FA9550-25-C-B010 and the Stanford Human-Centered AI Institute Hoffman-Yee grant program.

\bibliographystyle{unsrt}
\bibliography{neurips_2026}
\newpage
\section*{\Large Supplementary for \\\model: Adaptive Planning with Interleaved Vision-Language Thoughts} 

\appendix

The appendix is organized as follows: In \appref{app:data}, we include details about the task suite and data generation pipeline. In \appref{app:model}, we describe the implementation of our model, including training and compute details. In \appref{app:baselines}, we provide details about the baseline methods and evaluation protocol. Additionally,  we provide additional experimental results and analysis in \appref{app:results} and \appref{app:analysis}.

\section{Task Suite and Datasets}
\label{app:data} 

This section describes the task suite and data generation pipeline, including scene sampling, reference plan construction, and reasoning trace generation. We then provide examples of the training data. Lastly, we provide the dataset statistics.

\subsection{Task Suite}
\label{app:data:task_families}

\begin{figure}
    \centering
    \includegraphics[width=0.95\linewidth]{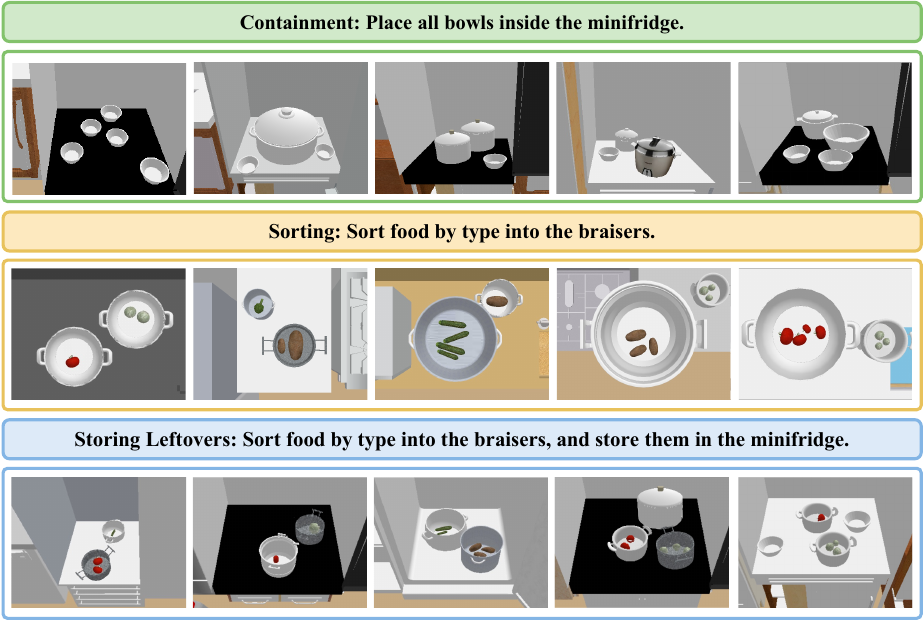}
    \captionsetup{labelfont=bf}
    \caption{KitchenWorlds task suite with increasing complexity. \textsc{Containment} (above) requires placing target objects into constrained storage regions, testing accessibility, obstruction handling, and free-space reasoning. \textsc{Sorting} (middle) requires semantic reasoning over object types to assign food objects to containers, while capacity constraints make some assignments geometrically infeasible. \textsc{Storing Leftovers} (below) composes both challenges into a longer-horizon held-out task, requiring the robot to sort food into containers and then store them in a constrained fridge.}
    \label{fig:dataset_examples}
\end{figure}

We define three task families: \textsc{Containment}, \textsc{Sorting}, and \textsc{Storing Leftovers}, as shown in \autoref{fig:dataset_examples}. These families test different combinations of constraints, including semantic preconditions, obstruction handling, geometric placement under limited free space.

\subsubsection{\textsc{Containment}.} 

\paragraph{Goal.} The goal of this task is to place a set of target objects (e.g. bowls) into a region (e.g. fridge or cabinet). The scene is initialized with all target objects visible on the counters, while the target region may vary in accessibility, obstruction, and available free space. 

\paragraph{Core Challenges.} The core challenges include recognizing semantic preconditions (e.g. door is open), resolving geometric constraints (e.g. remove the distractor object to make space for the target objects), and planning precisely if geometrically constrained (e.g. precise target object placement). Thus, harder instances require reasoning about accessibility, obstruction, and placement feasibility.

\paragraph{Variations.} We vary the number and sizes of target objects, the door state of the container, whether the target region is empty or obstructed, the number of distractor objects, and the amount of available free space. This produces instances ranging from simple open-container cases with ample space to constrained cases requiring prerequisite actions, obstacle removal, and precise placement.

\subsubsection{\textsc{Sorting}.} 

\paragraph{Goal.} The goal is to sort a set of food objects by type into separate target containers. Each instance contains two food types (e.g., zucchini and tomato) and two containers (e.g., braisers), and the robot must place all objects of each type into a separate container. The scene is initialized with all target objects and containers visible on the counter, so the core challenge is semantic assignment under capacity constraints. 

\paragraph{Core Challenges.} A valid assignment of food types to containers depends on the object geometries and container capacities. To verify whether an assignment is valid, we sample placements in the simulator to check whether all objects can be placed collision-free inside their assignment containers. We define two assignment regimes: both-valid and one-valid. In the both-valid regime, either type can fit into either container. In the one-valid regime, one type exceeds the capacity of the smaller container and only one assignment succeeds, requiring a unique mapping from food type to container. Thus, harder instances cannot be solved by semantic grouping alone. They require the planner to perform assignment based on capacity and placement feasibility.

\paragraph{Variations.} To induce these challenges, we vary the number and sizes of each object type and the relative capacities of the containers. This produces samples for both regimes that range in how visually obvious the assignment is.

\subsubsection{\textsc{Storing Leftovers.}}
\paragraph{Goal.} The goal is to store a set of food objects as leftovers by first sorting them into containers and then placing the containers in the fridge. Each instance contains two food types, two containers (e.g., braisers), and a minifridge. The robot must place all objects of each food type into a separate container, and then store the filled containers in the fridge. This task combines the capacity-based assignment challenge from \textsc{Sorting} with the access and free-space constraints from \textsc{Containment}. We hold out this task family from training to evaluate whether models can generalize from the component families to a novel compositional task.

\paragraph{Core Challenges.} A valid plan depends on both the food-to-container assignment and the feasibility of storing the resulting containers in the fridge. As in Sorting, the assignment of food types to containers may depend on object geometries and container capacities. In addition, the final fridge placement may require opening the fridge, clearing obstacles, and arranging the containers within limited free space. Thus, harder instances require reasoning about how early assignment and placement decisions affect downstream storage feasibility.

\paragraph{Variations.} As in the \textsc{Sorting} and \textsc{Containment} tasks, we vary the number and sizes of food objects, the relative capacities of the containers, the fridge door state, and fridge interior. This produces longer-horizon tasks that require composing semantic grouping, capacity reasoning, and geometric placement constraints. 

\newpage

\subsection{Data Generation Pipeline}
\label{app:data:pipeline} 
We generate all training and evaluation data in KitchenWorlds~\cite{yang_guiding_2024}, a PyBullet-based simulation environment. The data generation pipeline consists of the following four stages:

\paragraph{Step 1: Scene sampling.} Given a task family, we sample a scene that includes the task-relevant object categories. Furthermore, we systematically vary the initial scene layout, including the number of task-relevant objects, distractors, and level of geometric constraint. These variations produce samples with varying levels of difficulty, from simple settings with ample free-space to harder cases requiring opening doors, removing obstacles, or carefully arranging objects under tight space. 

\paragraph{Step 2: Successful reference plan generation.} To generate a reference plan for a  sampled task instance, we use PDDLStream~\cite{garrett_pddlstream_2020} with FastDownward~\cite{helmert_fast_2006} to generate a feasible executable plan, and then convert the resulting feasible executable trajectory into the compact action space predicted by our model.

If a successful trajectory is found, we process the PDDLStream output into a reference plan. The original output contains both high-level actions and low-level continuous parameters, such as object poses, grasps, robot configurations, and motion trajectories. Since our model predicts grounded high-level actions rather than low-level robot trajectories, we filter the PDDLStream plan to retain the actions corresponding to \texttt{open}, \texttt{pick}, and \texttt{place}. Specifically, \texttt{open} actions are parameterized by the articulated object, \texttt{pick} primitives are parameterized by the object to grasp, and \texttt{place} primitives are parameterized by the object being placed, the target support surface, and a normalized image-space placement point obtained from the grounded placement pose. This yields a successful, simulator-validated reference plan consisting of grounded high-level actions.

In addition, we save the simulator state, symbolic plan information, and low-level command sequence. These saved artifacts allow us to replay the plan, compute per-action metadata, and render RGB observations used to construct reference outputs needed for supervised training.

\paragraph{Step 3: Structured skeleton construction.} 
Given the reference plan, our goal is to construct an interleaved language-vision reasoning trace for supervised finetuning. To do so, we first extract a structured skeleton for this trace. 

First, we decompose the reference plan into a sequence of subgoals. Concretely, \texttt{open} actions become an \emph{open} subgoal, and continuous \texttt{pick-place} actions are grouped into a single \emph{place} subgoal. Place subgoals are further categorized by whether the object is being placed into the goal receptacle or moved elsewhere, such as when clearing an obstacle or repositioning an object. This sequence of subgoals captures the structure of the reference plan while abstracting away action-level details.

Next, each subgoal is annotated with a purpose, constraint type, and modality label. Possible subgoal purposes include advancing the task, satisfying semantic precondition, and clearing away obstacles. To determine the constraint type and modality label, we use simulator-derived metadata obtained by replaying the saved plan and saving geometric metrics. For place subgoals, we compute the available free space on the target surface, the free-space margin for the current object, and the remaining free space after placement. We also estimate future feasibility by comparing the remaining free space after a placement to the footprint of the remaining goal objects that still need to be placed. These quantities are computed from object bounding boxes projected onto the X-Y plane. Although these metrics are approximate, they provide a consistent simulator-grounded signal for identifying subgoals that depend on spatial constraints. 

Based on these annotations, each subgoal is labeled with a constraint type: symbolic precondition, current geometric constraint, or future geometric feasibility. We use these constraint annotations to assign a modality label to each reasoning step. Steps with a symbolic precondition constraint are assigned text-only reasoning. For steps with either a current geometric or future geometric feasibility constraint, we assign it to use an image based on whether the geometric heuristic reaches a pre-defined threshold.

\paragraph{Step 4: Reference trace expansion.}
After constructing the annotated skeleton trace, we use a high-capacity language model to expand it into a natural-language reasoning trace. The skeleton specifies the sequence of decision points, the corresponding actions, each decision's purpose, its primary planning constraint, the assigned modality, and any associated visual reference. The LLM is instructed to preserve this structure while adding connective reasoning about task decomposition, subgoal selection, progress toward the goal, relevant physical constraints, and why language alone or a visual thought is appropriate for each step. The generated reasoning is then parsed back into a structured format, so that the final reference output contains both the natural-language reasoning trace and the simulator-validated target plan. We use \texttt{gemini-3-flash-preview} for this step. The trace expansion prompt is provided in Appendix~\ref{app:prompts:expansion}.

For each reasoning step assigned a visual modality, we insert a visual thought after the text component of the step. A visual thought is represented in the trace by a discrete visual-token span \texttt{<|image\_start|><|image\_pad|>\ldots<|image\_pad|><|image\_end|>}, and is paired with a rendered RGB observation after the subgoal has been achieved. During training, the rendered image provides the latent target for the visual thought tokens.

\subsection{Dataset Examples}
\label{app:reasoning_samples}
\begin{figure}
    \centering
    \includegraphics[width=0.95\linewidth]{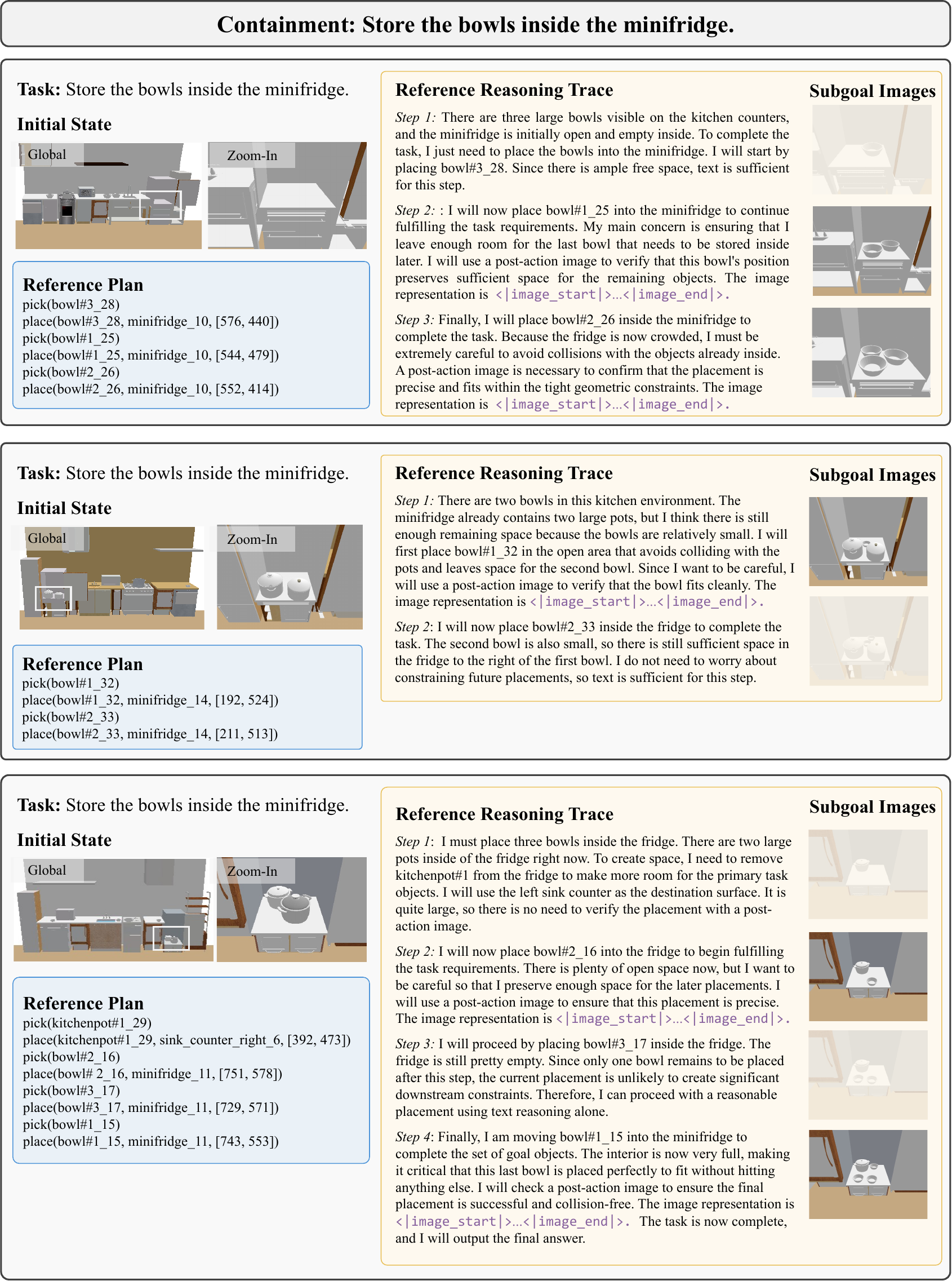}
    \captionsetup{labelfont=bf}
    \caption{Training examples for \textsc{Containment}, showing the inputs  and reference outputs.}
    \label{fig:containment_train_examples}
\end{figure}

\begin{figure}
    \centering
    \includegraphics[width=0.95\linewidth]{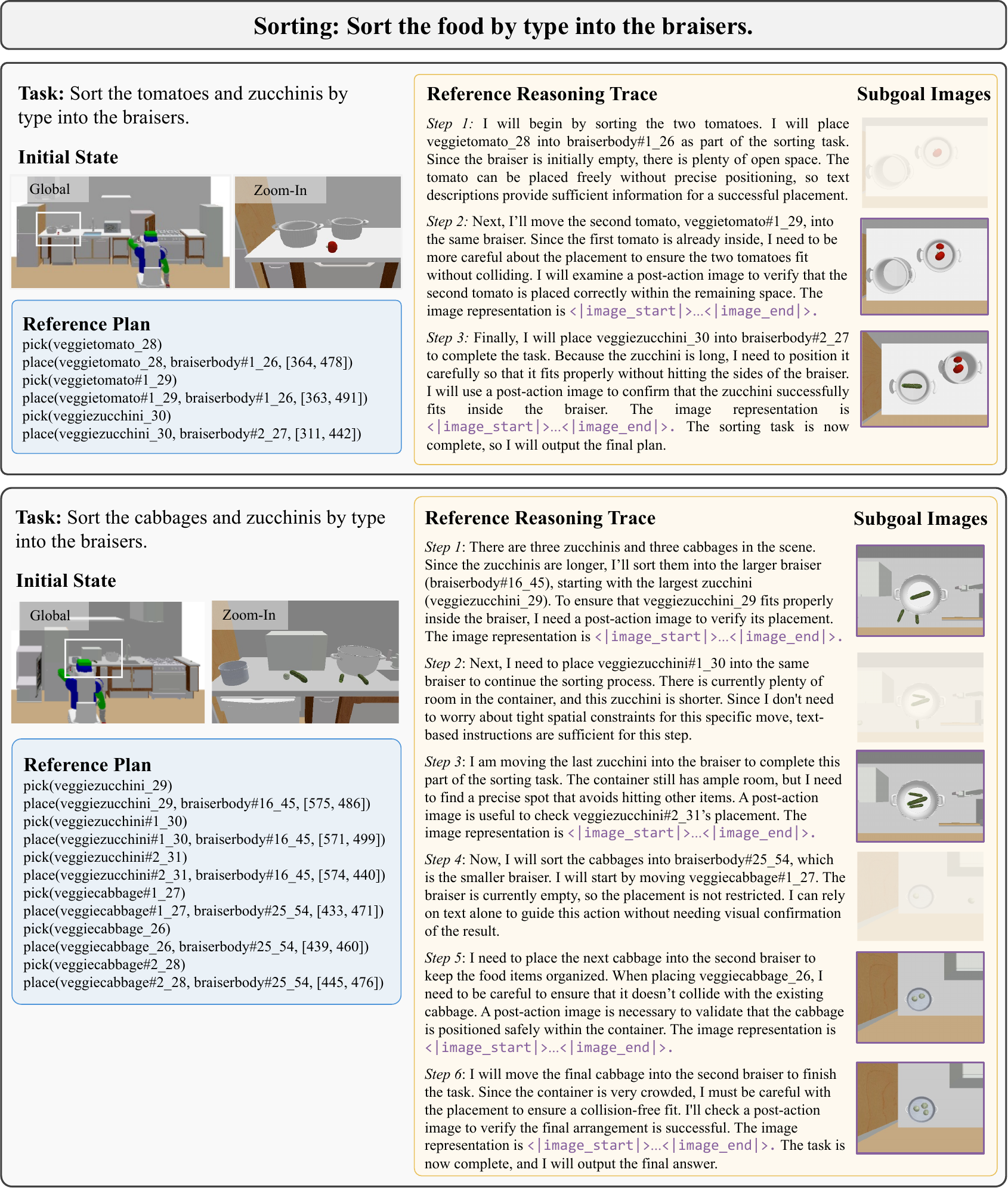}
    \captionsetup{labelfont=bf}
    \caption{Training examples for \textsc{Sorting}, showing the inputs  and reference outputs.}
    \label{fig:sorting_train_examples}
\end{figure}

In \autoref{fig:containment_train_examples} and \autoref{fig:sorting_train_examples}, we provide dataset examples for each training task family. These demonstrate the task instruction and input scene, alongside the reference outputs (reasoning trace, subgoal images, and plan). 

\subsection{Dataset Statistics} 
Using the above pipeline, we construct datasets for each training stage and evaluation. For each training stage, the dataset contains 2{,}000 examples from each training task families. \textsc{Containment} instances contain 1--4 target objects, and \textsc{Sorting} instances contain 2--5 target objects. We also systematically vary the presence of distractor objects, scene layouts, and initial states.

For evaluation, each dataset contains 100 examples. We include two in-distribution (ID) datasets, one for each training task family, sampled from the same distribution of target objects and constraints as the training data. We also evaluate on the held-out compositional task family, \textsc{Storing Leftovers}, with instances containing 2-5 target objects for sorting. Finally, we construct two within-family out-of-distribution (OOD) datasets to evaluate generalization to longer task horizons and increased task complexity. For task horizon, we increase it to 5--7 target objects for \textsc{Containment} and 6--10 for \textsc{Sorting}. For complexity, we increase the number of prerequisite actions required, defined based on actions that do not directly contribute to the goal (e.g. removing obstacles, opening doors).

\newpage 

\section{Model and Training Details}
\label{app:model} 

We provide implementation details for \model, including the architecture used to represent latent visual thoughts and the training setup used for finetuning.

\subsection{Architecture and Visual Thought Implementation}
Our model is initialized from Qwen3-VL-8B-Instruct and extends the base VLM decoder to support latent visual thoughts within the autoregressive reasoning trace. We represent each visual thought as a fixed-length span of special tokens: 
$\texttt{<|image\_start|>}
\;\texttt{<|image\_pad|>}^{K}\;
\texttt{<|image\_end|>}$,
where $K=16$. Since Qwen3-VL-8B has hidden dimension $d=4096$, each visual thought corresponds to a latent visual state $H_j \in \mathbb{R}^{16 \times 4096}.$

The start and end tokens of the visual-thought span supervise the model to produce valid visual-thought structure, while the hidden states corresponding to the image-pad tokens represent the model's latent visual thought.

\subsection{Training and Compute}
\label{app:model:training_and_compute}

We finetune \model with LoRA applied to the language-model attention and MLP projection modules, with rank $r=16$ and scaling $\alpha=32$.

We train our model with the three-stage supervised fine-tuning curriculum described in Section~\ref{sec:method:training}. We train on 2000 examples for each task family, for 1 epoch. Across all training stages, we use a per-device batch size 4 with gradient accumulation over 16 steps, optimizer AdamW, cosine learning-rate decay, learning rate $4\times10^{-5}$, and a warmup ratio $0.08$. 

For the training objective: In Stage 1, we set $\lambda_{\mathrm{CE,trace}}=1.0$ and $\lambda_{\mathrm{vis}}=0$. In Stages 2 and 3, we set $\lambda_{\mathrm{CE,trace}}=1.0.$ and $\lambda_{\mathrm{vis}}=5.0$.

Training is performed on a single RTX PRO 6000 GPU with 96GB of memory, with each run taking approximately 14 hours. 

\newpage

\section{Evaluation: Execution Protocol and Baselines}
\label{app:baselines}

In this section, we describe the closed-loop execution protocol used to evaluate all models, followed by details of the general-purpose VLM and planning baselines.

\subsection{Closed-Loop Execution Protocol}
\label{app:baselines:execution_protocol}

We evaluate all methods in the same receding-horizon execution setup, using our simulator. 

Formally, the environment is modeled as a tuple $(\mathcal{S}, \mathcal{A}, \mathcal{T})$, where $\mathcal{S}$ is the state space, $\mathcal{A}_{\mathrm{sim}}$ is the simulator action space, and $\mathcal{T}:\mathcal{S}\times\mathcal{A}_\mathrm{sim}\rightarrow\mathcal{S}$ is the transition function. In our setup, $\mathcal{A}_{\mathrm{sim}}$ consists of parameterized action primitives (e.g., \texttt{open}, \texttt{pick}, and \texttt{place}) whose continuous parameters (e.g., grasp pose, object pose) are defined in the world frame. 

The planner does not directly output actions in $\mathcal{A}_{\mathrm{sim}}$. Instead, it predicts actions in an analogous action space $\mathcal{A}_{\mathrm{plan}}$ that preserves the same discrete primitives but only expresses placement parameters. Specifically, as described in Section~\ref{sec:method}, the planner action space $\mathcal{A}_{\mathrm{plan}}$ consists of three high-level primitives:
\texttt{open(obj)}, \texttt{pick(obj)}, and \texttt{place(obj, target, u, v)}. The \texttt{open} action opens articulated containers such as cabinets or refrigerators. The \texttt{pick} action selects an object instance, and \texttt{place} places the held object on or inside a target object using an image-space placement coordinate $(u,v)$ normalized to the $[0,1000]$ range.  

During execution, planner actions are mapped into $\mathcal{A}_{\mathrm{sim}}$ by lifting any image-space coordinates into world-frame using the simulator state, known camera pose, and scene geometry. Concretely, we unproject the predicted image coordinate into the scene, raycast against the simulator geometry, and place the object at the resulting 3D point. The simulator then applies the corresponding primitive and updates the state according to $\mathcal{T}$. 

Episodes terminate on task success,  or when the maximum number of execution steps or wall-clock time is reached. Task success is defined as satisfying all goal predicates at termination. 

\subsection{General-Purpose VLM Baselines}
\label{app:baselines:vlm}

We evaluate general-purpose VLMs by prompting them directly to produce plans in the same action format as our model. Each prompt contains the current image observation, the task goal, the object list with image-space object locations, and the action schema (See Figure~\ref{fig:text_only_prompt}). The model outputs a plan, which is parsed and executed using the shared closed-loop protocol above. 

For Gemini-based baselines, we evaluate Gemini-3.1-Pro and Gemini-ER-1.5 using greedy decoding with temperature $0$. For Qwen baselines, we evaluate Qwen3-VL-8B-Instruct and Qwen3-VL-8B-Thinking. Extended thinking is enabled for Gemini-3.1-Pro, Gemini-ER-1.5, and Qwen3-VL-8B-Thinking. Their thinking tokens are included in token usage measurements.

\subsection{Planning Baselines}
\label{app:baselines:planning}

\paragraph{FastDownward.}
We evaluate FastDownward~\cite{helmert_fast_2006} as a symbolic planning baseline. Since FastDownward cannot operate directly on RGB observations, we train BLADE~\cite{liu2024learning} on our generated dataset to infer the symbolic predicates required by the planner from the image and language input. The inferred state is converted into a PDDL problem and solved with FastDownward using A* search with the LM-cut heuristic.  Because FastDownward produces symbolic actions rather than image-space placement coordinates, we ground its placement actions with random sampling.

\paragraph{Reflect-VLM.}
We evaluate Reflect-VLM~\cite{feng_reflective_2025} as a VLM planning baseline that uses external, imagined future observations to critique and revise its plans before execution. To adapt Reflect-VLM to our task setting, we train it on our generated planning dataset using the same task instances and action space as our method. At inference, Reflect-VLM receives the current image, goal, and object locations, and proposes a plan with $n=5$ high-level actions, following their implementation. The proposed plan is rolled out using a diffusion-based dynamics model for a fixed number of imagined steps, producing future observations that are given to the VLM for reflection. 

\paragraph{VLM-TAMP.}
VLM-TAMP~\cite{yang_guiding_2024} is a representative VLM-based task-and-motion planning framework that uses a VLM to propose symbolic planning structure and a downstream planner to produce executable actions. In this framework, the VLM receives the same input as the general-purpose VLM baselines, namely the current image, goal, object locations, and action schema, and is prompted to propose a sequence of formal subgoals. These subgoals are then solved by a symbolic planner and grounded using the same execution interface.

\newpage 
\section{Additional Results}
\label{app:results}

In this section, we provide additional out-of-distribution generalization results and ablations of the main components of \model's training and reasoning pipeline. Concretely, we analyze the role of domain-specific finetuning, the contribution of each training stage, the effect of reasoning modality, and the impact of heuristic-guided reasoning trace generation.

\subsection{Out-of-Distribution Generalization}

\paragraph{Experiment Setup.} We further evaluate \model{} on out-of-distribution (OOD) settings that increase task horizon and geometric complexity relative to the training distribution. The \textsc{Longer Horizon} setting increases the number of target objects, while the \textsc{Increased Complexity} setting includes tighter free-space constraints, and additional prerequisite steps such as opening storage regions or removing obstacles before placement. 

\paragraph{Results.} As shown in \autoref{tab:ood_gen}, \model{} remains competitive across these harder settings, achieving comparable or stronger performance than the general-purpose VLMs and planning baselines. This suggests that its learned use of visual thoughts can transfer to more complex scenes, although the performance margin is smaller than in-distribution (ID).

However, \model{} shows one of the largest ID-OOD drops compared to baselines. The OOD settings contain more objects than those seen in training, requiring visual thoughts to encode denser spatial configurations. Thus, we hypothesize that this drop in performance is partly due to limited supervision for generating visual thoughts of more cluttered scenes.  

\label{app:results_ood_gen} 
\begin{table}[h]
\captionsetup{labelfont=bf}
  \caption{Within-family OOD generalization. We report the success rate and relative ID-OOD gap.}
  \label{tab:ood_gen}
  \centering
  \small
  \setlength{\tabcolsep}{4pt}
  \resizebox{\linewidth}{!}{%
  \begin{tabular}{l cc cc}
    \toprule
     & \multicolumn{2}{c}{Success Rate $\uparrow$} & \multicolumn{2}{c}{Relative Gap $\downarrow$} \\
    \cmidrule(lr){2-3} \cmidrule(lr){4-5}
    Model & Long Horizon & More Complex & Long Horizon & More Complex \\
    \midrule
    \multicolumn{5}{l}{\textit{VLM Baselines}} \\
    Gemini-3.1-Pro       & 0.225 & 0.224 & 0.082 & 0.086 \\
    Gemini-ER-1.5        & 0.331 & 0.303 & \textbf{0.021} & 0.104 \\
    Qwen3-VL-8B-Instruct & 0.177 & 0.188 & 0.059 & 0.000 \\
    Qwen3-VL-8B-Thinking & 0.199 & 0.235 & 0.142 & \textbf{-0.013} \\
    \midrule
    \multicolumn{5}{l}{\textit{Planning Baselines}} \\
    FastDownward  & 0.213 & 0.243 & 0.217 & 0.107 \\
    Reflect-VLM   & 0.242 & 0.252 & 0.171 & 0.137 \\
    VLM-TAMP      & 0.281 & 0.269 & 0.146 & 0.182 \\
    \midrule
    \textbf{\model{} (Ours)} & \textbf{0.338} & \textbf{0.336} & 0.193 & 0.199 \\
    \bottomrule
  \end{tabular}%
  }
\end{table}

\subsection{Text-Finetuning Ablation}
\label{app:ablations:text_finetune}

\paragraph{Experiment Setup.} To isolate whether \model's gains come from domain-specific finetuning alone, we train a text-only variant using the same Qwen3-VL-8B-Instruct backbone, task instances, and expert plans as \model. This variant is trained with standard supervised finetuning on the text reasoning trace and final plan, providing a control for whether the model can recover the same planning improvements simply by learning task-specific language rationales and action formats.

\begin{figure}[h]
\centering
\includegraphics[width=0.55\linewidth]{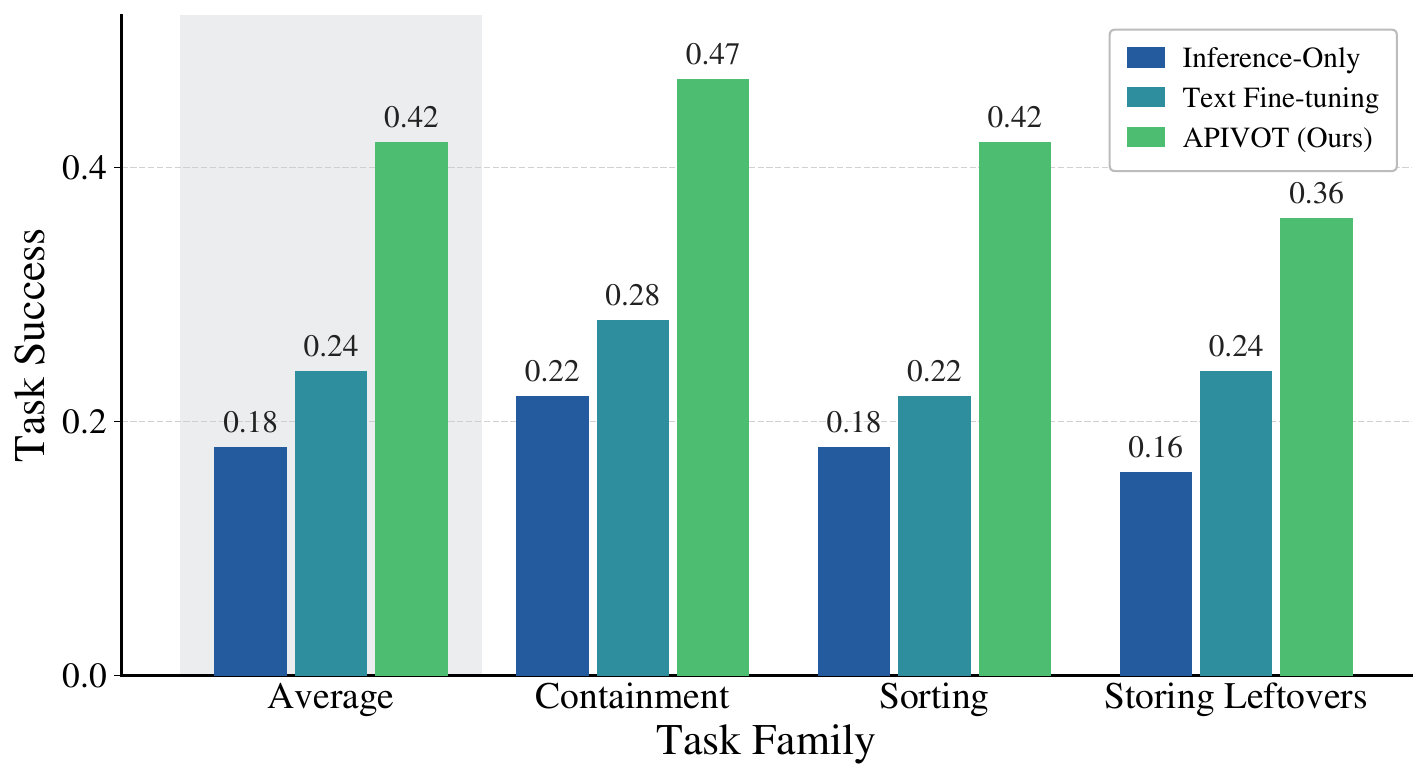}
\captionsetup{labelfont=bf}
\caption{Success rates across all task families.}
\label{fig:text_ablation}
\end{figure}

\paragraph{Results.} As shown in \autoref{fig:text_ablation}, text-only SFT improves over the base Qwen3-VL-8B-Instruct model, increasing average task success from $0.188$ to $0.24$. This indicates that domain-specific finetuning is useful. The model benefits from exposure to the task distribution, structured reasoning format, and expert action sequences. However, text-only SFT yields limited improvement in comparison to \model, which achieves an average success rate of $0.419$. Thus, text-only SFT accounts for only a small fraction of the overall gain. The remaining gap suggests that \model's improvement is not simply due to better task-specific language supervision, but to its ability to represent intermediate future states visually and leverage them for planning.

\subsection{Training Stage Ablation}

\label{app:ablations:training_stages}

\paragraph{Experiment Setup.} \model is trained through a three-stage curriculum that progressively teaches it to understand, generate, and adaptively use  visual thoughts for planning (Section~\ref{sec:method:training}). To evaluate the contribution of each stage, we ablate one stage at a time starting from the full \model training pipeline and evaluate on all tasks. This isolates the contribution of each training stage in learning effective and efficient planning.

\paragraph{Results.} We report the average success rate and token usage for each ablation in \autoref{tab:training_ablation}. We see that removing Stage 1 substantially reduces average success by $11.8$ points. This suggests that the model does not learn to leverage visual thoughts for planning from later visual generation supervision alone. Without Stage 2, \model's performance decreases by $7.4$ points, indicating that learning to generate visual-thought at every step provides an important prior for generation behavior. Without this stage, \model does not reliably instantiate visual thoughts when needed, as supported by the decrease in token count. Lastly, removing Stage 3 yields the highest success rate, but also increases token count substantially. This highlights Stage 3's role in teaching the model to invoke visual thoughts selectively, while maintaining performance. Without this stage, the model relies on visual thoughts uniformly, which can improve success but at a substantially higher reasoning cost. 

Overall, we see that Stages 1 and 2 are crucial for learning effective visual-thought representations, while Stage 3 helps to achieve a favorable performance-cost balance. 

\begin{table}[h]
    \centering
    \captionsetup{labelfont=bf}
    \caption{Ablation of APIVOT's three-stage training curriculum.}
    
    \label{tab:training_ablation}
    \begin{tabular}{l cc}
        \toprule
        Method & Success Rate & Token Usage \\
        \midrule
        \textbf{\model} & \textbf{0.419} & \textbf{262} \\
        \midrule
        \quad w/o Stage 1: Comprehension & 0.301 & 235 \\
        \quad w/o Stage 2: Generation & 0.345 & 206 \\
        \quad w/o Stage 3: Adaptation & 0.457 & 498 \\
        \bottomrule
    \end{tabular}
\end{table}

\subsection{Reasoning Modality Ablation}
\label{app:ablations/reasoning_modality}
\paragraph{Experiment Setup.} To disentangle the effects of training-time multimodal supervision and inference-time multimodal generation, we ablate the modality used at training and inference. We compare \model against a text-only variant finetuned on text-only traces, as in Appendix~\ref{app:ablations:text_finetune}. For \model, we further ablate the reasoning modality at inference-time: 1) text-only, where visual thoughts are not used, 2) always-image, where images are generated at every step, 3) adaptive, our learned policy that balances performance and cost, and 4) GT oracle, where \model conditions on ground-truth visual thoughts as an upper bound. 

\paragraph{Results.} As shown in \autoref{tab:thought_modality_ablation}, text+image training improves planning performance over text-only finetuning overall. Even when inference time reasoning is text-only, the text+image model outperforms the text-only model by $3.8$ percentage points. This suggests that multimodal supervision provides benefits beyond explicit visual generation at test time, likely by encouraging planning representations that are more sensitive to geometry. 

Inference-time visual thoughts provide an even larger gain, as demonstrated by the performance of the image-always and adaptive ablations. Finally, oracle visual thoughts achieve the highest average success rate, outperforming generated visual thoughts across all task families. This gap suggests that improving the quality of generated visual thoughts could yield additional gains.

\begin{table}[h]

\captionsetup{labelfont=bf}

  \caption{Effect of training and inference modality on planning performance.}

  \label{tab:thought_modality_ablation}

  \centering

  \definecolor{ourrow}{gray}{0.92} 

  \begin{tabular}{l l c c c c}

    \toprule

    Training & Inference

    & Avg

    & \textsc{Contain}

    & \textsc{Sort}

    & \textsc{Store} \\

    \midrule

    Text & Text & 0.244 & 0.281 & 0.217 & 0.235 \\

    \midrule

    \multirow{4}{*}{Text+Image (\model)}

       & Text                       & 0.282 & 0.332 & 0.267 & 0.246 \\

       & Image (always)             & 0.459 & 0.483 & 0.466 & 0.428 \\

       & \textbf{Image (adaptive)}          & \textbf{0.419} & \textbf{0.472} & \textbf{0.421} & \textbf{0.365} \\

       & Image (GT, oracle) & 0.482 & 0.494 & 0.482 & 0.470 \\

    \bottomrule

  \end{tabular}

\end{table}

\subsection{Heuristic-Guided Reasoning Trace Ablation}

\label{app:ablations:heuristic}

\paragraph{Experiment Setup.} \model's third SFT stage teaches adaptive modality selection by training on reasoning traces where visual thoughts are inserted selectively for geometrically constrained subgoals. As described in Section~\ref{sec:method:supervision} and Appendix~\ref{app:data:pipeline}, these placements are determined based on simulator-derived heuristics. To isolate the effect of this placement strategy, we initialize all models from the same Stage 2 checkpoint and vary the dataset used for Stage 3 training. We compare our final heuristic-based dataset, constructed to use visual-thoughts $75\%$ of the time, against a random baseline that inserts visual thoughts independently at each step with probability $0.75$. This controls for the effectiveness of the geometric-heuristic. We also evaluate heuristic datasets with lower usage rates of $50\%$ and $25\%$.

\paragraph{Results.} As shown in \autoref{tab:heuristic_ablation}, the heuristic-based $75\%$ configuration achieves the best success rate. Although the Random baseline uses more visual thoughts at inference-time, it performs worse than our heuristic strategy. This suggests that the geometric heuristic provides an effective signal for learning to use visual thoughts adaptively when beneficial.

However, reducing the visual-thought rate to $50\%$ and $25\%$ degrades performance. Visual-thought usage appears to collapse at inference-time, with models using them on only $20\%$ and $3\%$ of steps. This suggests that sparse visual-thought supervision weakens the model's learned ability to use visual representations, even when initialized from a Stage 2 checkpoint that generated visual thoughts at every step. These results suggest the need for more robust strategies for learning adaptive modality selection.

\begin{table}[h]
    \centering
    \captionsetup{labelfont=bf}
    \caption{Reasoning trace heuristic ablation. \textbf{Bold} is the setting used to train \model.}
    \label{tab:heuristic_ablation}
    \begin{tabular}{l c cc}
        \toprule
        \multicolumn{2}{c}{Reference Traces} & \multicolumn{2}{c}{Inference-Time} \\
        \cmidrule(lr){1-2} \cmidrule(lr){3-4}
        Modality Strategy & Visual Thought Usage (\%) & Success Rate & Visual Thought Usage (\%) \\
        \midrule
        \textbf{Heuristic (Ours)} & \textbf{75} & \textbf{0.419} & \textbf{56} \\
        Heuristic & 50 & 0.332 & 20 \\
        Heuristic & 25 & 0.254 & 3  \\
        \midrule
        Random & 75 & 0.371 & 63 \\
        \bottomrule
    \end{tabular}
\end{table}

\newpage
\section{Additional Analysis}
\label{app:analysis} 

\subsection{Qualitative Examples}
\label{app:qualitative_examples}
We provide qualitative comparisons against the top-performing VLM (Gemini-ER-1.5) and planning baseline (VLM-TAMP) across all three tasks: \textsc{Containment}, \textsc{Sorting}, and \textsc{Storing Leftovers}. Across these examples, \model more reliably anticipates spatial constraints and produces physically feasible placements, leading to success while baselines struggle. 

\newpage

\begin{figure}[H]
    \centering
      \captionsetup{labelfont=bf}
      \includegraphics[width=\linewidth]{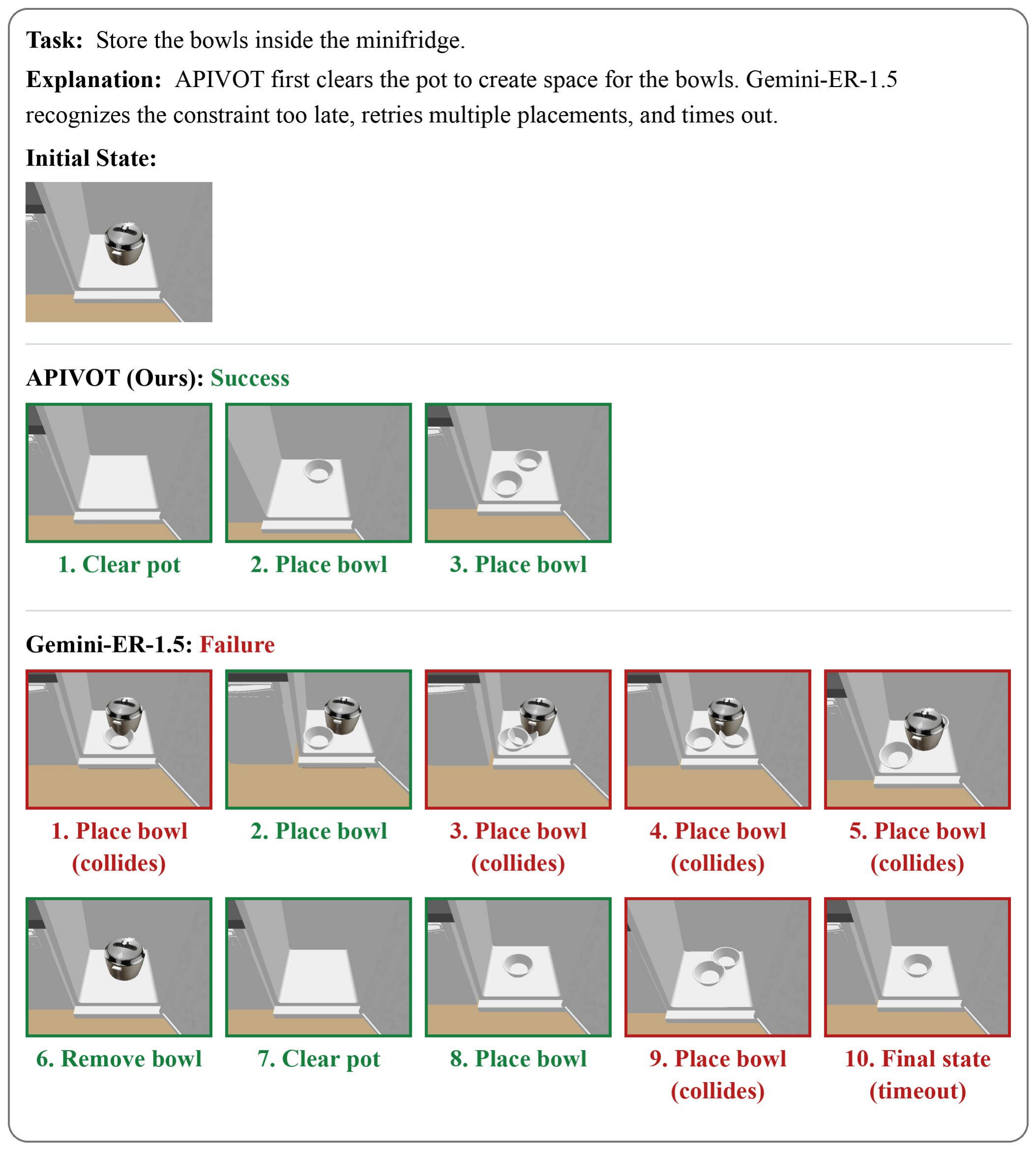}
        \caption{Qualitative comparison of \textsc{Containment}: \model clears an obstacle to complete the task, while Gemini-ER-1.5 recognizes the space constraint too late and times out.}

  \centering
  \vspace{-0.5cm} 
  
\label{fig:app:containment_1} 

\end{figure}

\begin{figure}[H]
    \centering
      \captionsetup{labelfont=bf}
      \includegraphics[width=\linewidth]{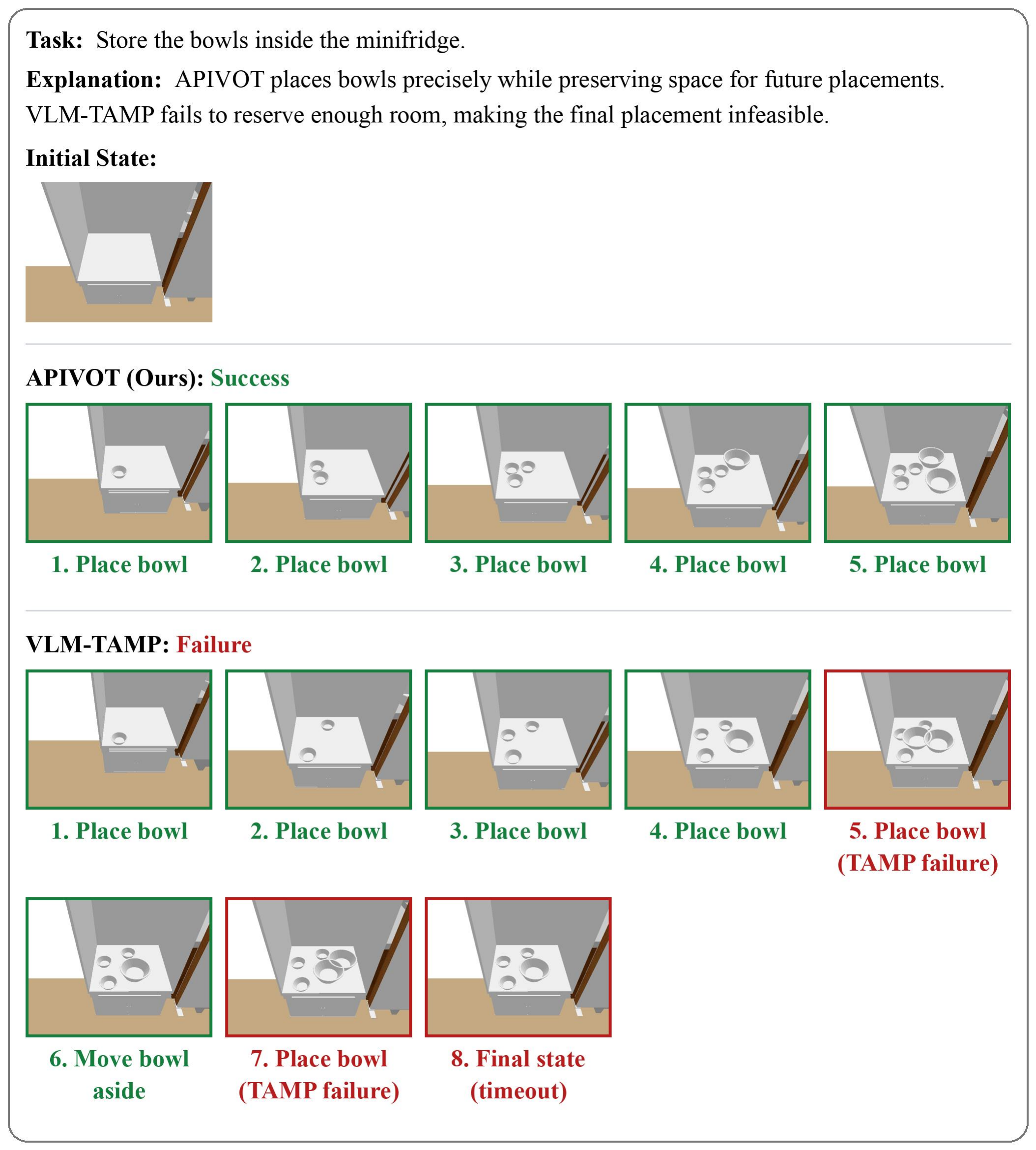}
    \caption{Qualitative comparison of \textsc{Containment}: \model preserves space for future bowl placements, while VLM-TAMP fails to anticipate the final placement constraint.}
  \centering
  \vspace{-0.5cm} 
  
\label{fig:app:containment_vlm_tamp} 
\end{figure}

\begin{figure}[H]
    \centering
      \captionsetup{labelfont=bf}
      \includegraphics[width=\linewidth]{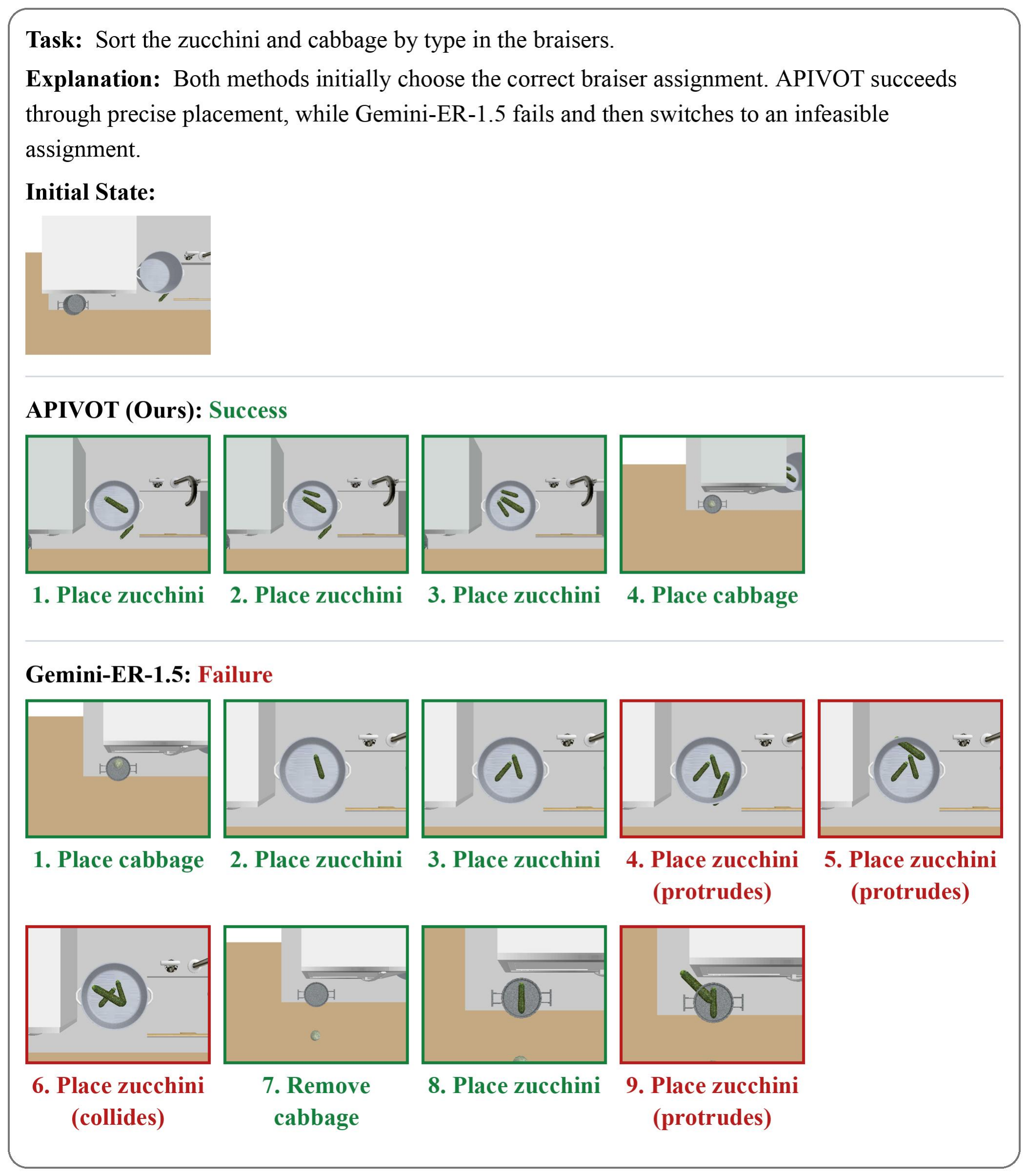}
        \caption{Qualitative comparison of \textsc{Sorting}: \model succeeds with the correct braiser assignment, while Gemini-ER-1.5 fails and switches to an infeasible assignment.}
  \centering
  \vspace{-0.5cm} 
  
\label{fig:app:sorting_gemini} 
\end{figure}

\begin{figure}[H]
    \centering
      \captionsetup{labelfont=bf}
      \includegraphics[width=\linewidth]{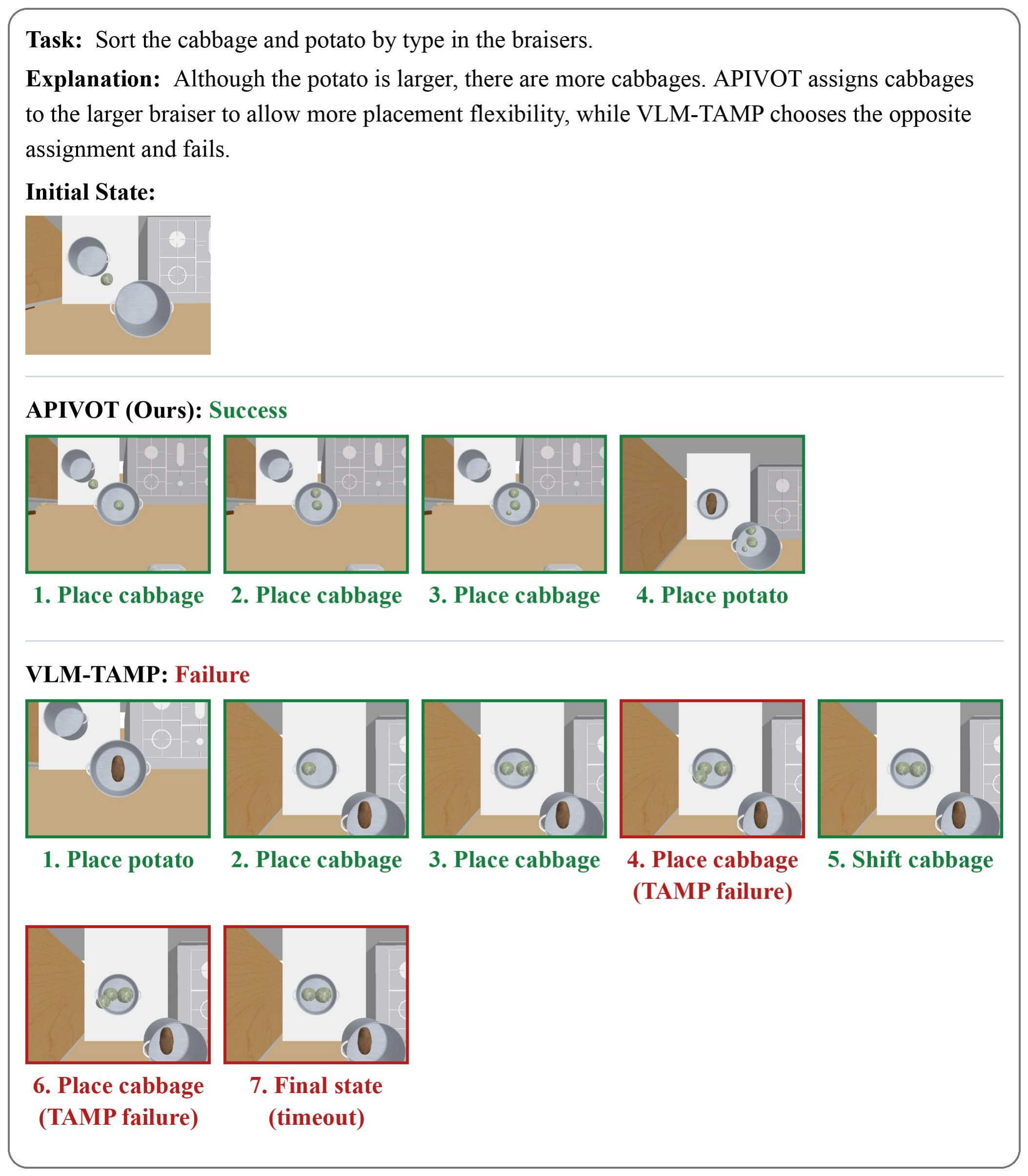}
        \caption{Qualitative comparison of \textsc{Sorting}: \model succeeds by selecting a more spatially flexible braiser assignment, while VLM-TAMP chooses the opposite, leaving insufficient room for placement.}
    \label{fig:app:sorting_vlm_tamp}
  \centering
  \vspace{-0.5cm}

\end{figure}

\begin{figure}[H]
    \centering
      \captionsetup{labelfont=bf}
      \includegraphics[width=\linewidth]{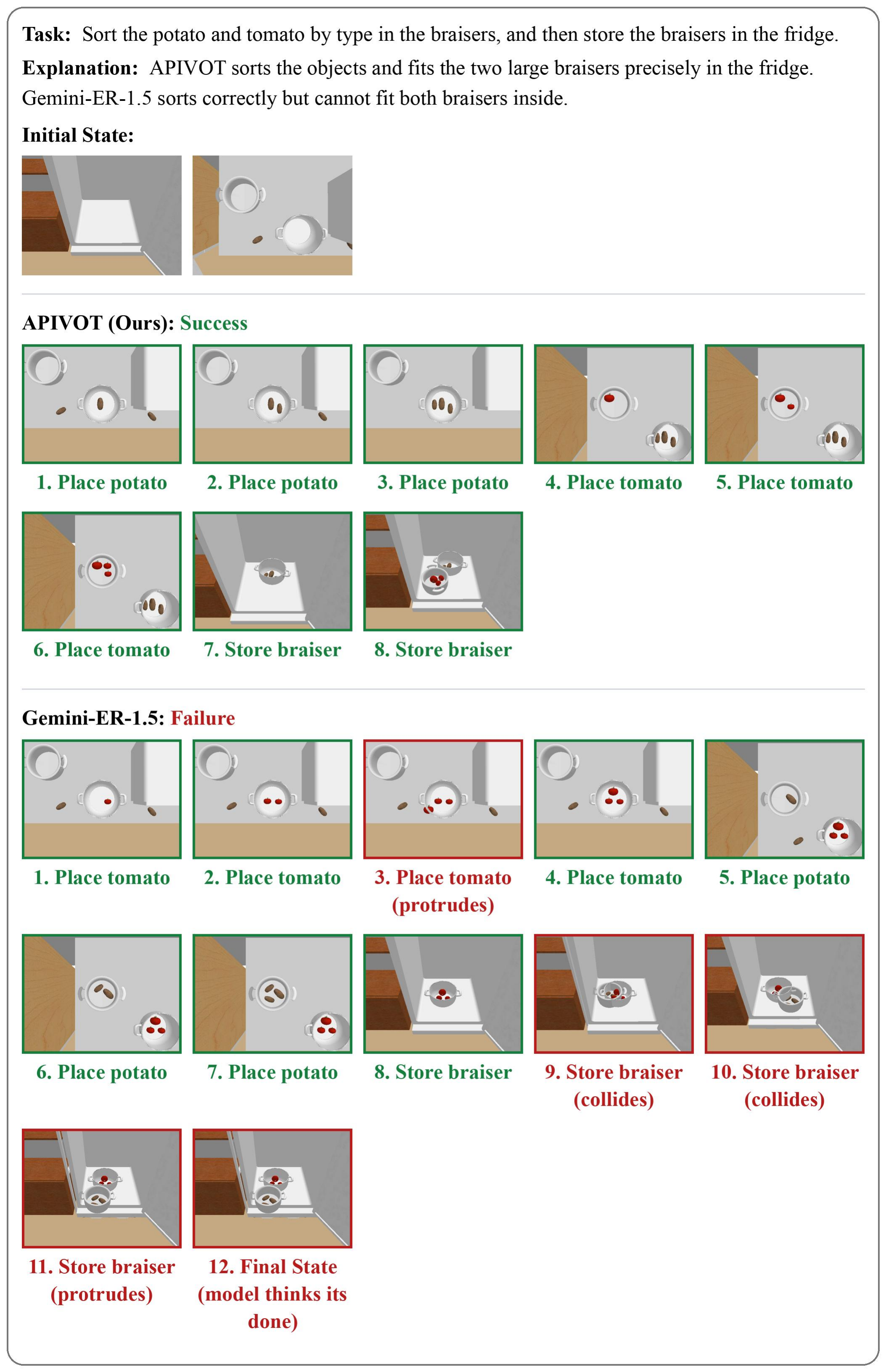}
        \caption{Qualitative comparison of \textsc{Storing Leftovers}: \model sorts the objects and fits both large braisers in the fridge, while Gemini-ER-1.5 cannot place both braisers feasibly.}

    \label{fig:app:storing_gemini}
  \centering
  \vspace{-0.5cm}
\end{figure}

\begin{figure}[H]
    \centering
      \captionsetup{labelfont=bf}
      \includegraphics[width=\linewidth]{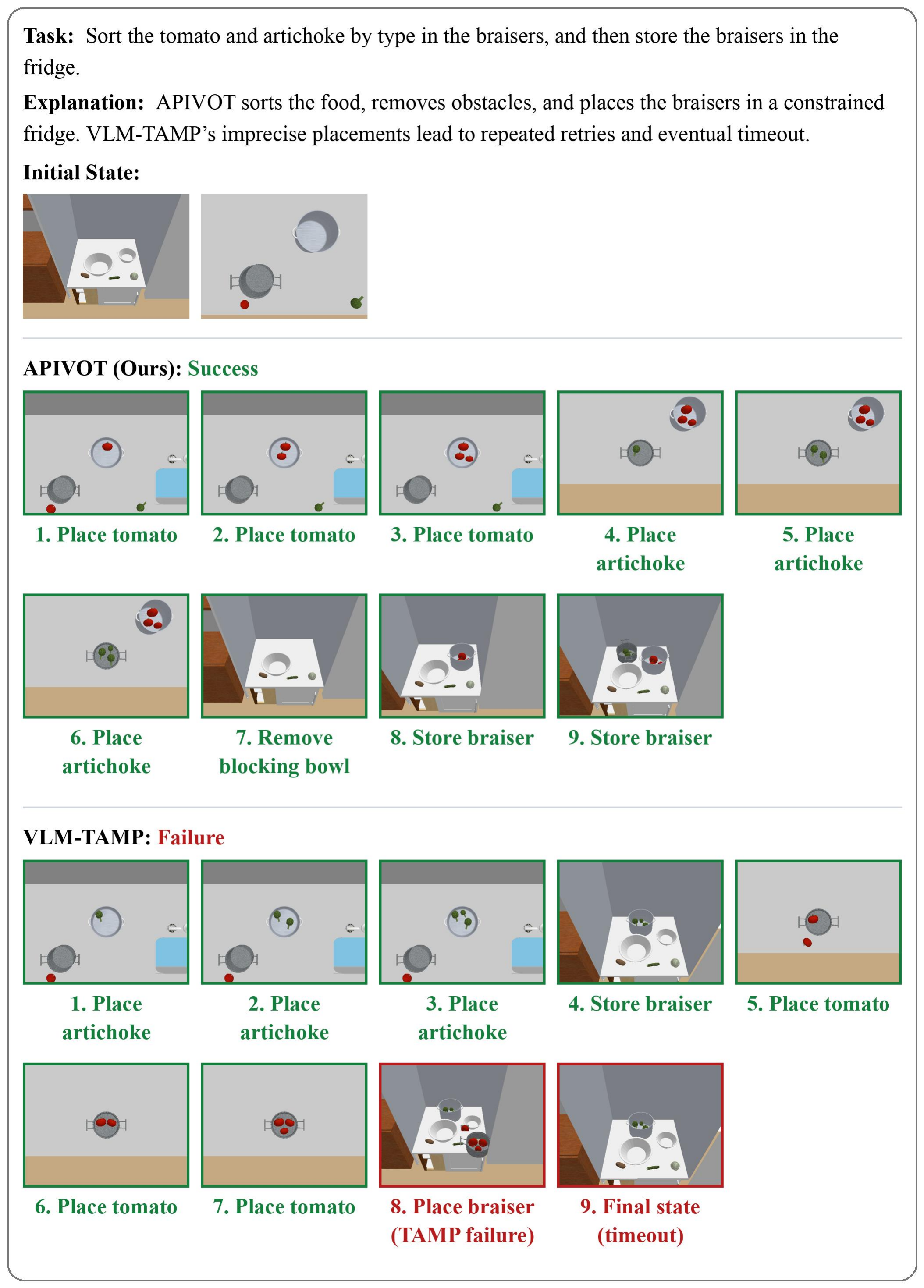}
        \caption{Qualitative comparison of \textsc{Storing Leftovers}: \model completes sorting, obstacle removal, and placement in a constrained region, while VLM-TAMP times out after several errors.}
    \label{fig:app:storing_vlm_tamp}
  \centering
  \vspace{-0.5cm}

\end{figure}

\newpage

\subsection{Failure Analysis}

We classify \model failures into the following primary failure modes:

\textit{Premature goal completion}: Since \model infers task progress from the current observation and object inventory, it sometimes misjudges a partially completed task as finished and consequently ends the episode without the remaining goal-relevant actions. We observe that this failure mode is more common on long-horizon tasks, suggesting the need for explicit progress tracking.

\textit{Incorrect action ordering}: The plan contains all necessary actions but orders them incorrectly. \model occasionally identifies the relevant objects and operations, but does not correctly model their temporal dependencies. For example, \model may plan to clear an obstacle after first placing an object there.

\textit{Missing prerequisite actions}: \model sometimes omits actions that are necessary to make the task solvable, yet not directly relevant to the goal. \model handles semantic prerequisites (e.g. opening a closed door) reliably, but still struggles on geometric constraints that require removing an obstacle or rearranging objects that block target locations, where the need for the action is apparent only from the spatial configuration of the scene.

\textit{Geometric grounding}: The high-level plan is executable in principle, but the system incorrectly localizes the placements. We identify these failures when the plan succeeds with oracle simulator placement but fails using \model’s predicted points. These failures are more common on tightly constrained scenes, where the tolerance for placement is smaller.

\newpage
\section{Prompt Templates}
\label{app:prompts}

Here, we provide the prompt templates for data generation and evaluation of VLM baselines.

\subsection{Data-Generation LLM Expansion Prompt}
\label{app:prompts:expansion}

We provide the prompt used to do trace expansion for the reference reasoning traces used for SFT. We use \texttt{gemini-3-flash-preview} for trace expansion.  

\begin{figure}[H]
\centering
\begin{tcolorbox}[
title={\small \textbf{Data-Generation Reasoning Prompt}},
width=1\columnwidth,
colback=gray!3,
colframe=gray!60,
colbacktitle=gray!15,
coltitle=black,
boxrule=0.5pt,
arc=1mm,
left=1mm,
right=1mm,
top=1mm,
bottom=1mm]
\fontsize{6pt}{7pt}\selectfont
\ttfamily

You are given structured planning steps from a robot task-planning trace. Your job is to convert each step into natural reasoning that explains how a planner would think about the decision.\\[1mm]

For EACH step, write 2-3 sentences of reasoning that follow this structure:\\
\hspace*{3mm}1. State what the action does in the context of the task.\\
\hspace*{3mm}2. Identify the key constraint that matters at this step.\\
\hspace*{3mm}3. Explain why the selected modality (text or text\_image\_post) is appropriate.\\[1mm]

At the end, indicating that the task is complete and that it will output the final answer. Do NOT add this earlier. The reasoning should sound like a planner thinking through the decision, not like metadata description.\\[1mm]

Reasoning style requirements:\\
\hspace*{3mm}- Write in natural reasoning style.\\
\hspace*{3mm}- Write reasoning in first-person planner voice (e.g. `I need to...').\\
\hspace*{3mm}- Do not simply restate the reason string. Expand it into natural reasoning.\\
\hspace*{3mm}- Keep reasoning aligned to each step.\\
\hspace*{3mm}- Be concise but explanatory.\\[2mm]

\textbf{Output format:}\\
Return the reasoning as JSON.\\[1mm] 
\mbox{[}\{"step\_index": 0, "reasoning": "reasoning paragraph"\},\\
\hspace*{3mm}...\\
\hspace*{3mm}\{"step\_index": n, "reasoning": "reasoning paragraph + closing sentence"\}\mbox{]}\\[1mm]

\textbf{Requirements:}\\
\hspace*{3mm}- Output ONLY valid JSON.\\
\hspace*{3mm}- Do not include explanations outside JSON.\\
\hspace*{3mm}- Do not include markdown.\\
\hspace*{3mm}- Do not include commentary.\\
\hspace*{3mm}- One entry per step.\\
\hspace*{3mm}- Step indices must match input.\\
\hspace*{3mm}- For the FINAL step only, append one additional closing sentence.\\[2mm]

\textbf{Example Input:}\\
Step 0:\\
Action: Move carrot into fridge\\
Action purpose: place a goal object into the target container\\
Constraint type: future\_geometric\_feasibility\\
Modality: Text\_image\_post\\
Why this modality: placement must preserve space for remaining objects\\[1mm]

Example Output:\\
\{"step\_index": 0, "reasoning": "I move the carrot into the fridge to place one of the goal objects into the target container. The key constraint is preserving enough room for the other vegetables. A post-action image is useful because I need to check whether the exact placement leaves space for future objects."\}\\

--------------------------------\\

Reminder of your goal: For each step, you are expanding a structured planning decision into natural reasoning that reflects how a task planner would think. For the FINAL STEP ONLY, append an additional closing sentence. The closing sentence must clearly indicate that the task is complete and that it will output the final answer. Do NOT add this to any earlier steps.\\
Your reasoning should:\\
\hspace*{3mm}- clearly connect the action to the task goal\\
\hspace*{3mm}- explain the key constraint affecting the decision\\
\hspace*{3mm}- justify why text or visual inspection is needed\\
\hspace*{3mm}- sound like deliberate decision-making, not data description\\[1mm]

Do not copy the template wording, list fields, repeat the reason string verbatim, invent new motivations, or add unnecessary verbosity. Focus on producing clear, natural, step-by-step planner reasoning that could realistically guide execution of the task.\\[1mm]

Now generate reasoning for the following steps.\\[2mm]
\{INSERT STEPS FOR A GIVEN TRACE\} \\

\normalfont
\end{tcolorbox}
\vspace{-3mm}
\captionsetup{labelfont=bf}
\caption{Prompt used to expand structured planning traces into natural-language reasoning traces.}
\vspace{-5mm}
\label{fig:data_generation_prompt}
\end{figure}

\newpage 

\subsection{VLM Inference Prompt}

We include the prompts used to evaluate planning for both our method (\autoref{fig:apivot_prompt}) and text-only VLM baselines (\autoref{fig:text_only_prompt}). 
In both prompts, the model is provided with the input observations, goal, object list, task constraints, and action primitives, and then prompted to output step-by-step reasoning, followed by a plan consisting of one formal action per line. 

The prompt for planning with \model specifies the output as a structured \texttt{<think>} trace, where each step specifies the planning intention, key constraint, modality decision, and justification, followed by a formal \texttt{<answer>} plan.  

\begin{figure}
\centering
\begin{tcolorbox}[
title={\small \textbf{APIVOT Planning Prompt}},
width=1\columnwidth,
colback=gray!3,
colframe=gray!60,
colbacktitle=gray!15,
coltitle=black,
boxrule=0.5pt,
arc=1mm,
left=1mm,
right=1mm,
top=1mm,
bottom=1mm]
\fontsize{7pt}{8pt}\selectfont
\ttfamily
\raggedright
\obeylines

You are given an input image of a kitchen world scene. Your task is to reason about the image and plan a sequence of actions that accomplishes the following goal:
``\{goal\}''.
Respond with actions in a formal language defined by the following primitive actions: \{set\_of\_actions\}.
Each line must contain exactly ONE action.

Scene Objects:
Objects referenced in the actions must come from the following list. Each name is formatted as `\textless object\_category\textgreater\_\textless instance\_id\textgreater'. Each object is paired with the position of its center in the format [x, y] (normalized to 0-1000):
\{object\_id\_to\_points\}

\{task\_constraints\}

Planning Guidelines:
You are a mobile robot with \{n\_arms\} arms. You must obey the following commonsense rules:
1. You must have at least one empty hand before you can pick up an object.
2. You can only take actions on objects listed above.
3. You should aim to achieve the goal with the minimum number of actions.

Answer Format:
You must output exactly two sections:
\textless think\textgreater...\textless /think\textgreater
\textless answer\textgreater...\textless /answer\textgreater

--------------------------------

1) \textless think\textgreater section:

In the \textless think\textgreater section, you must produce a structured planning trace as an ordered sequence of reasoning steps written in the format:

Step k: ...

Each step should describe the next planning subgoal required to complete the task. Each step should include reasoning about the following:

1. Planning intention: State what must be decided or done next and how it contributes to achieving the goal.

2. Key constraint:
Identify the most important constraint affecting this step. Examples include:
- spatial capacity
- object accessibility
- ordering dependencies
- none

3. Modality decision:
Decide whether this step requires visual verification or whether text reasoning is sufficient.

4. Justification:
Explain WHY that modality choice is appropriate.

--------------------------------

Image representation rule:

If visual verification is required for a step, you must emit an image representation immediately after the reasoning for that step.

Correct structure example:

Step 2: I need to determine whether the container still has enough space for the remaining goal objects. The key constraint is preserving sufficient free space for future placements. I need a post-action image to verify that the remaining layout will leave enough free space for the other objects to fit. \textless IMAGE\_START\textgreater\textless IMAGE\_PAD\textgreater...\textless IMAGE\_END\textgreater

--------------------------------

2) \textless answer\textgreater section:

Output the final plan as a list of primitive actions.

Requirements:
- One action per line
- Use only the formal primitive action language
- Do not include reasoning here
- The plan must be consistent with the reasoning in \textless think\textgreater

--------------------------------

Final rules:

- Your output must follow EXACTLY this structure:
\textless think\textgreater...\textless /think\textgreater
\textless answer\textgreater...\textless /answer\textgreater

- Do NOT include anything outside these sections.
- Do NOT include explanations outside \textless think\textgreater.
- Do NOT include reasoning inside \textless answer\textgreater.
- Ensure the reasoning and plan are consistent.

\end{tcolorbox}
\captionsetup{labelfont=bf}
\caption{Planning prompt used for APIVOT.}
\label{fig:apivot_prompt}
\end{figure}

\begin{figure}
\centering
\begin{tcolorbox}[
title={\small \textbf{Text-Only VLM Evaluation Prompt}},
width=1\columnwidth,
colback=gray!3,
colframe=gray!60,
colbacktitle=gray!15,
coltitle=black,
boxrule=0.5pt,
arc=1mm,
left=1mm,
right=1mm,
top=1mm,
bottom=1mm]
\fontsize{6pt}{7pt}\selectfont
\ttfamily
\raggedright
\obeylines

You are given an input image of a kitchen world scene. Your task is to reason about the image and plan a sequence of actions that accomplishes the following goal:
``\{goal\}''.
Respond with actions in a formal language defined by the following primitive actions: \{set\_of\_actions\}.
Each line must contain exactly ONE action.

Scene Objects:
Objects referenced in the actions must come from the following list. Each name is formatted as `\textless object\_category\textgreater\_\textless instance\_id\textgreater'. Each object is paired with the position of its center in the format [x, y] (normalized to 0-1000):
\{object\_id\_to\_points\}

\{task\_constraints\}

Planning Guidelines:
You are a mobile robot with \{n\_arms\} arms. You must obey the following commonsense rules:
1. You must have at least one empty hand before you can pick up an object.
2. You can only take actions on objects listed above.
3. You should aim to achieve the goal with the minimum number of actions.

Output Format:
First, write step-by-step reasoning about how to complete the task, including task decomposition, subgoal setting, and action sequencing. Each step should focus on planning the next planning subgoal required to complete the task, and reasoning about what must be decided or done next and how it contributes to achieving the goal. 

Then, write "Answer:" on its own line, followed by the final plan as a list of actions with one action per line.

\end{tcolorbox}
\captionsetup{labelfont=bf}
\caption{Text-only planning prompt used for VLM baselines.}
\label{fig:text_only_prompt}
\end{figure}

\newpage


\newpage

\end{document}